\pdfoutput=1

\documentclass[11pt]{article}

\usepackage[]{ACL}

\usepackage{times}
\usepackage{latexsym}

\usepackage[T1]{fontenc}

\usepackage[utf8]{inputenc}

\usepackage{microtype}

\usepackage{inconsolata}

\usepackage{longtable}
\usepackage{xcolor}
\usepackage{colortbl}
\usepackage{multicol}
\usepackage{booktabs}
\usepackage{makecell}
\usepackage{amsmath, amssymb}
\usepackage{multirow}
\usepackage{mathtools}
\usepackage{enumitem}
\usepackage{subcaption}
\usepackage{float}
\usepackage{graphicx}
\usepackage{caption} 
\usepackage{upgreek}
\usepackage{seqsplit}
\usepackage{color,soul}
\usepackage{arydshln}
\usepackage{xspace}

\usepackage{algorithm}
\usepackage{algorithmic}

\usepackage{amsmath,amsfonts,bm}

\def\eqref#1{equation~\ref{#1}}

\def\1{\bm{1}}

\DeclareMathAlphabet{\mathsfit}{\encodingdefault}{\sfdefault}{m}{sl}
\SetMathAlphabet{\mathsfit}{bold}{\encodingdefault}{\sfdefault}{bx}{n}

\usepackage[nameinlink]{cleveref}

\hypersetup{
    colorlinks = true,
    citecolor = mountainmeadow,
    linkcolor = crimson,
    linktoc = all,
    urlcolor = darkblue,
}

\newcommand{\highlight}[1]{{\color{crimson}{#1}}}

\newcommand{\relp}{Rel.\xspace} 
\newcommand{\ours}{COINCIDE\xspace}

\makeatletter

\newcommand{\COMMENTARR}[1]{%
  \leavevmode\hfill$\triangleright$~#1}

\newcommand\blfootnote[1]{%
  \begingroup
  \renewcommand\thefootnote{}\footnote{#1}%
  \addtocounter{footnote}{-1}%
  \endgroup
}

\definecolor{Blue}{rgb}{0,0,0.8}
\definecolor{mountainmeadow}{rgb}{0.19, 0.73, 0.56}
\definecolor{crimson}{rgb}{0.86, 0.08, 0.24}
\definecolor{darkblue}{rgb}{0.0, 0.0, 0.55}
\definecolor{ggr}{gray}{0.92}
\definecolor{azure(colorwheel)}{rgb}{0.0, 0.5, 1.0}

\definecolor{gg}{HTML}{E0FEFE}

\newcolumntype{a}{>{\columncolor{ggr}}c}

\title{Concept-skill Transferability-based Data Selection\\ for Large Vision-Language Models}

\author{
    Jaewoo Lee$^{1}$ \; 
    Boyang Li$^{\dagger,2}$ \; 
    Sung Ju Hwang$^{\dagger,1,3}$ \\
    KAIST$^{1}$ Nanyang Technological University, Singapore$^{2}$ DeepAuto$^{3}$ \\
    \texttt{jwlee8877@gmail.com boyang.li@ntu.edu.sg sjhwang82@kaist.ac.kr}
}

\begin{document}
\maketitle

\begin{abstract}
Instruction tuning, or supervised finetuning on extensive task-specific data, is necessary for Large Vision-Language Models (LVLMs) to generalize well across a broad range of vision-language (VL) tasks. However, training on large VL datasets can become prohibitively expensive. In this work, we introduce \ours, an effective and scalable data selection technique that uses a small model as a reference model to select visual instruction tuning data for efficient finetuning of a target LVLM, focusing on diversity and transferability. Specifically, we cluster the training data using internal activations from a small model, which identifies VL concept-skill compositions needed by a target LVLM. We then sample data from these diverse clusters by considering their density and transferability, or the ability to transfer well to other concept-skill compositions. This approach ensures the diversity of these compositions, which is vital for LVLM generalization. Extensive experiments demonstrate that \ours achieves superior performance and data selection efficiency against 8 strong baselines on two distinct datasets: LLaVA-1.5 and Vision-Flan. Using only 20\% of the LLaVA-1.5 dataset, \ours achieves performance comparable to the LVLM finetuned on the whole dataset, with 70\% reduction of the wall-clock running time. On the Vision-Flan dataset, our method achieves superior results with only 16.7\% of the training data. Our code is available at \href{https://github.com/G-JWLee/COINCIDE_code}{\textcolor{magenta}{https://github.com/G-JWLee/COINCIDE\_code}}. \blfootnote{$\dagger$ Equal advising}
\end{abstract}

\section{Introduction}
Large Vision-Language Models (LVLMs)~\citep{Zhu2023minigpt4, Dai2023instructblip, Radford2021clip, Zhai2023siglip} are often built by (1) pretraining on paired image-caption datasets and (2) subsequent finetuning on image-instruction data on diverse vision-language (VL) tasks. The second step, referred to as visual instruction tuning (VIT), substantially enhances multimodal instruction-following capabilities. To achieve broad generalization, recent works~\citep{Cha2023honeybee, Dong2024internlm4k, Chen2024internvl15, Li2024minigemini} integrate an increasing number of VL tasks into VIT. 

However, training on extensive VIT data incurs significant computational cost, making the process infeasible for small academic labs and individual researchers. Additionally, it is not clear if all the VIT data are necessary for good generalization, as different VL tasks have different abilities to transfer to downstream tasks \cite{Tiong2024lvlmfactoranal, Xi2023transferability, Ostapenko2024modularllm}. 

\begin{figure}[t]
    \centering
    \includegraphics[width=\linewidth]{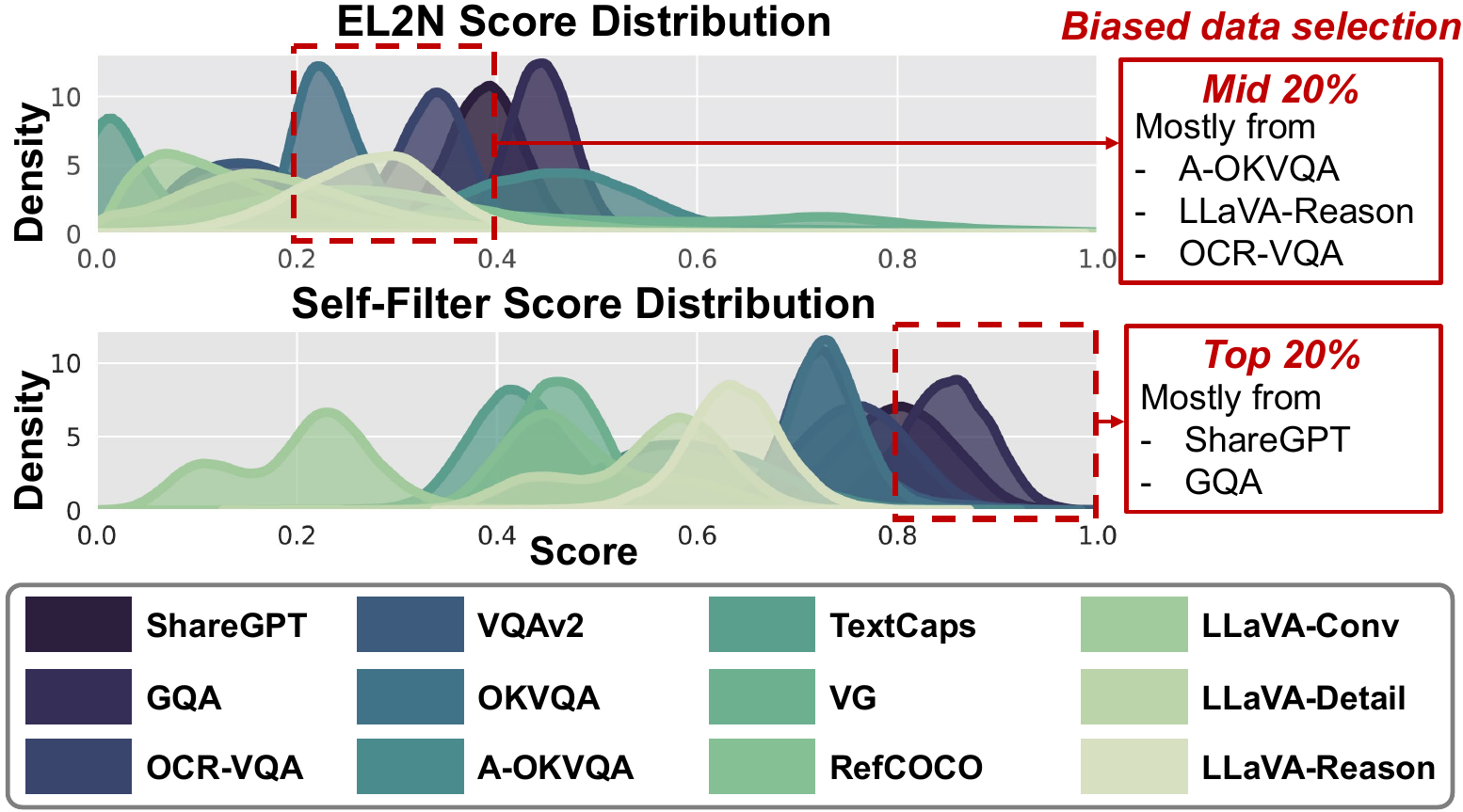}
    \par
    \caption{Different VL tasks in LLaVA-1.5~\citep{Liu2023llava15} exhibit different score distributions. Thus, selecting data based on a single score metric like EL2N~\citep{Paul2021el2n} or Self-Filter~\citep{Chen2024selffilter} results in a biased coreset (\textcolor{red}{red}), substantially decreasing the diversity within the coreset.}
    \vspace{-0.1in}
    \label{fig:score_distribution}
\end{figure}

In this paper, we investigate the selection of a coreset, a subset that approximates the performance of the full dataset, from large VIT datasets. Conventional coreset selection approaches~\citep{Marion2023llmpre, Zhou2023lima, Chen2023alpagasus} usually utilize a score metric to select training data. As VIT datasets are highly diverse and feature multiple data modes (\Cref{fig:score_distribution}), data selection using any single metric would produce a coreset dominated by a few tasks. \Cref{fig:score_distribution} indicates that, selecting 20\% of data from any part of the metric distribution of EL2N~\citep{Paul2021el2n} or Self-Filter~\citep{Chen2024selffilter} would exclude many data modes, which severely reduces the diversity of the selected coreset and harms generalization. As our experiments show (\Cref{tab:llava_eval}), this type of coreset selection degrades LVLM performance.

\begin{figure}[t]
    \centering
    \includegraphics[width=\linewidth]{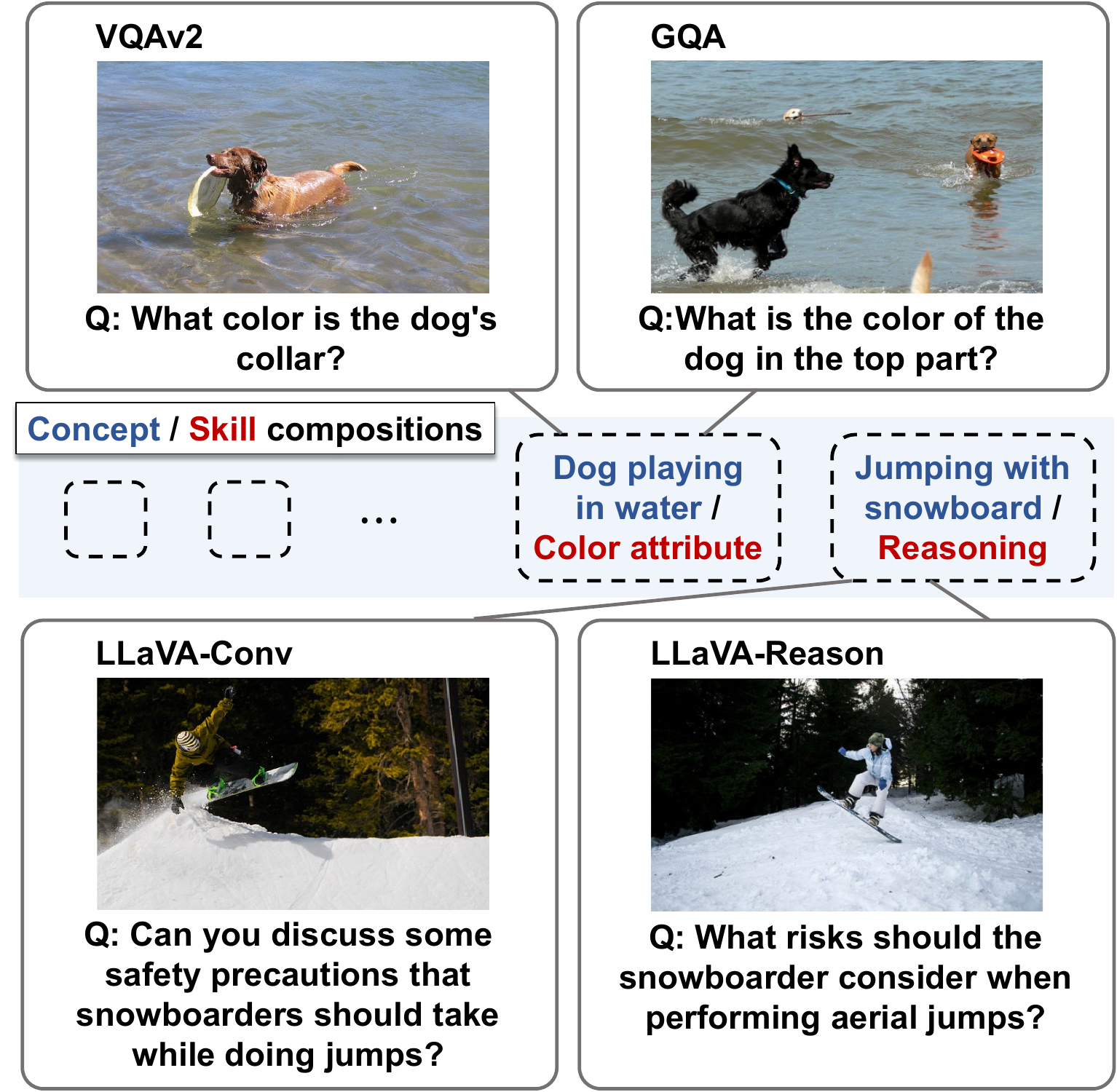}
    \par
    \caption{Different VL tasks (e.g., VQAv2 and GQA, LLaVA-Conv and LLaVA-Reason) share VL concept-skill compositions.}
    \label{fig:shared_vl_concept_skill}
\end{figure}

Our solution to the multitude of data modes is straightforward: we explicitly identify the modes by clustering the VIT data points using features from multiple layers in a small LVLM. Interestingly, we observe that the clusters thus identified roughly coincide with compositions of VL concepts and skills.
For example, a concept could be street signs or trains on a railroad, while a skill could be OCR, recognizing color, or reasoning. Upon close inspection, we find that different VL tasks contain overlap over these concept-skill compositions. As exemplified in~\Cref{fig:shared_vl_concept_skill}, LLaVA-Conv and LLaVA-Reason contain questions about the risks of snowboard jumps, despite their separate focuses on multi-turn conversations and reasoning. This suggests sampling over the clusters would be more effective in enhancing the diversity of VL concept-skill compositions than sampling over datasets or tasks. 

\begin{figure*}[t]
    \centering
    \includegraphics[width=\linewidth]{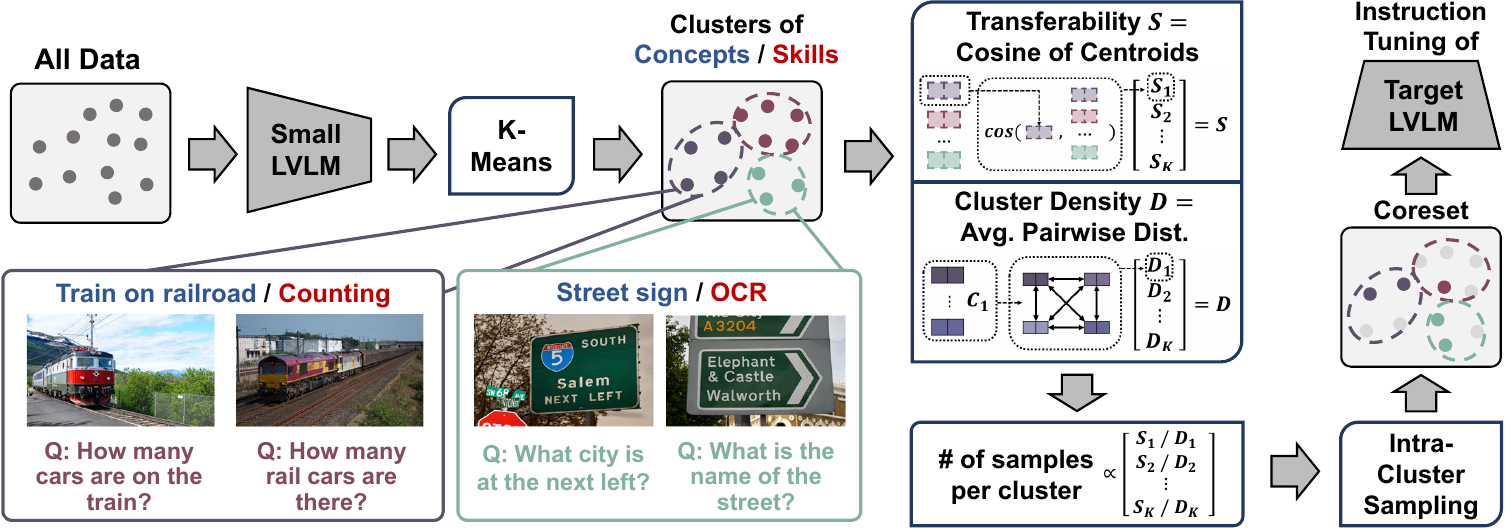}
    \par
    \caption{Illustration of \ours. Our method utilizes a small LVLM to cluster visual instruction tuning data based on concept-skill compositions. We then assess the cluster transferability as the mean cosine similarity to other cluster centroids. We further compute the cluster density as the mean Gaussian kernel distance among all data pairs within the cluster. Using cluster transferability and density, \ours determines the number of data to sample from each cluster and performs intra-cluster sampling. Finally, it combines all the selected samples from all the clusters to compose the final coreset.}
    \label{fig:overview}
\end{figure*}

To this end, we introduce \textbf{CO}re \textbf{IN}struction \textbf{C}oncept-sk\textbf{I}ll \textbf{D}ata \textbf{E}lection (\ours), which identifies VL concept-skill compositions through data clustering using activations from an off-the-shelf, small LVLM (\Cref{fig:overview} \highlight{Left}). From each cluster, \ours selects training data for a target LVLM by considering transferability (i.e., how well knowledge from each cluster can facilitate LVLM's learning in other clusters) and internal density of clusters (\Cref{fig:overview} \highlight{Right}). Empirically, we find that transferability correlates well with cosine similarity among clusters. Based on the findings, we select more data points from more transferable clusters. Further, we sample fewer data points from denser clusters, as data points in dense clusters are likely redundant. 

Another major challenge of coreset selection is its high computational cost. Existing techniques often require expensive steps like additional training~\citep{Du2023MoDS, Mekala2024small, Chen2024selffilter}, gradient calculation~\citep{Xia2024less, Liu2024tive}, or the use of bigger and more advanced models~\citep{Chen2023alpagasus, Liu2023deita}. The time complexity and the assumption of larger models contradict the primary goal of coreset selection, which is to reduce the development cost of new models larger than existing ones. In comparison, \ours assumes only a VLM (2B) smaller than the target LVLM (7B, 13B) and does not require any backward pass.

We validate the effectiveness of \ours across a wide range of coreset selection scenarios using two distinct VIT datasets, LLaVA-1.5~\citep{Liu2023llava15} and Vision-Flan~\citep{Xu2024visionflan}. The experimental results demonstrate that our method achieves performance competitive with that of the LVLM finetuned with the full dataset, with 30\% of time cost including the data selection and training. Our approach also achieves superior performance and efficiency compared to 8 strong baselines.

In summary, our contributions are as follows:
\begin{itemize}[itemsep=0.0mm, parsep=1pt]
  \item We introduce \ours, an efficient coreset selection pipeline for a target LVLM using an existing small reference model to cluster training data. Training on 16.7-20\% data selected by \ours achieves comparable performance to whole-dataset finetuning, leading to 70\% wall-clock time reduction.
  \item We propose an efficient transferability calculation among clusters based on our novel observation of a positive correlation between cluster centroid similarity and cluster transferability.
  \item To enhance training efficacy, we prioritize samples from clusters with high transferability and low density, while still selecting a few samples from other clusters for diversity.
\end{itemize}

\section{Related Work}
\label{sec:related_work}
\paragraph{Coreset Selection} Coreset selection attempts to extract a subset of training data that functions comparably to the full training set. This technique is adopted for problems like active learning~\citep{Wei2015activecoreset1, Sener2018activecoreset2}, continual learning~\citep{Rebuffi2017clcoreset1, Aljundi2019clcoreset2}, and data pruning~\citep{Pleiss2020aum, Paul2021el2n}. Recent works~\citep{Zhou2023lima, Xia2024less}  investigate coreset selection for instruction tuning of LLMs. Alpagasus~\citep{Chen2023alpagasus} uses ChatGPT~\citep{ChatGPT} to rate the quality of instruction samples. S2L~\citep{Yang2024s2l} leverages the training loss trajectory of smaller models to find optimal samples for training larger LLMs. DiverseEvol~\citep{Wu2023diverseevol} utilizes the target model itself to iteratively choose beneficial data for the current training episode.

\paragraph{Coreset Selection for Visual Instruction Tuning}
Several very recent papers address the coreset selection problem for visual instruction tuning~\citep{Wei2023instructgpt4, Chen2024selffilter, Liu2024tive}. Self-Filter~\citep{Chen2024selffilter} scores VIT data using a score-net trained along with the target LVLM. The concurrent work TIVE~\citep{Liu2024tive} employs gradient information from the target LVLM to compute task- and sample-level importance. Although effective, it demands considerable memory to store the high-dimensional gradient vectors. Moreover, these methods require backward passes, which are expensive due to the large training set. Both also overlook the diversity of selected data, which is vital for generalization. In contrast, our approach reduces wall-clock running time and considers both transferability and diversity.

\paragraph{VL Concept and Skill Discovery}
Concept discovery in neural networks is a key topic in interpretability research (\citealt{pmlr-v80-kim18d, NEURIPS2023_abf3682c, Manning2020}). Notably, \citet{kowal2024visual} performs hierarchical clustering in layer-wise activation space. \citet{Tiong2024lvlmfactoranal} attempts to identify latent skills underlying VL datasets. \citet{Michaud2023llmskill} performs spectral clustering to discover LLMs skills. Though these works provide inspiration, they are orthogonal to our work, whose main objective is to sample from data clusters rather than understanding existing neural networks. The only application of concept discovery we are aware of is by \citet{Gupta_2017_ICCV}, showing consistent VL concepts improve transfer learning.

\section{Method}
We start by introducing the framework that utilizes neuron activations from a small LVLM to group VIT data into clusters, where each cluster comprises samples exhibiting a similar concept-skill composition (\Cref{sec:subsec:clustering}). Next, we conduct experiments to examine the correlation between the similarity of a cluster centroid to other centroids and the transferability of that cluster to others (\Cref{sec:subsec:correlation_discover}). Based on our findings, we describe our data selection strategy, which performs cluster-wise sample selection by selecting different numbers of samples from clusters depending on their transferability and diversity (\Cref{sec:subsec:our_approach}). The overall framework of our approach is illustrated in~\Cref{fig:overview}.

\subsection{Preliminaries \label{sec:subsec:preliminaries}}

A modern LVLM typically consists of a visual encoder and an LLM, which are connected by intermediate network layers. The visual information is fed to the LLM as input (\citealt{Dai2023instructblip, Liu2023llava}), or guides cross-attention (\citealt{alayrac2022flamingo}). Here we focus on a transformer-based LLM that receives visual information as input tokens. 

The $l$-th transformer layer receives the visual tokens $\bm{x}^{v}_{l} \in \mathbb{R}^{N_v \times D}$ and text tokens $\bm{x}^{t}_{l} \in \mathbb{R}^{N_t \times D}$, where $N_v$ and $N_t$ are the numbers of tokens, and $D$ is the hidden dimension size. A transformer layer contains a multi-head self-attention (MSA) and a feed-forward network (FFN). For the purpose of this paper, we describe only MSA formally:
\begin{equation}
    \hspace*{-2mm}[\bm{z}^{v}_{l},\bm{z}^{t}_{l}]=\texttt{MSA}_{l}\left(\texttt{LN}_{l}\left([\bm{x}^{v}_{l},\bm{x}^{t}_{l}]\right)\right)+[\bm{x}^{v}_{l},\bm{x}^{t}_{l}],
    \label{eq:tokens_processing}
    \hspace*{-2mm}
\end{equation}
where $[\cdot,\cdot]$ denotes concatenation, $\texttt{LN}_{l}$ denotes layer normalization, and $\bm{z}^{v}_{l}$ and $\bm{z}^{t}_{l}$ are output visual and text features from the $l$-th layer MSA, respectively.

\subsection{Discovering Concept-Skill Compositions \label{sec:subsec:clustering}}
An LVLM aims to learn about a large variety of visual-linguistic concepts and skills. Hence, it is important to automatically sort training data into concepts and skills, so that the coreset can provide sufficient coverage of these. Recent studies~\citep{Schwettmann2023multimodalneuron1, Pan2023multimodalneuron2, Gandelsman2024clipneuron} reveal that the internal activations at various layers of LVLMs may encode different visual concepts.

To figure out which layer of the LVLM provides the best feature representation for visual concept and skill discovery, we perform a preliminary visualization study of TinyLLaVA-2B~\citep{Zhou2024tinyllava}. Given an image and a textual question, we visualize the image patches that contribute the most to the generation of the ground-truth answer. Using features from different layers highlights different image patches. Ideally, we can compare the visualization with human intuition and select the layer that agrees with human intuition the most. We provide detailed experimental procedures with some visualization results in~\Cref{appendix:relevancy_map}.

Perhaps surprisingly, we find that the best layer varies substantially according to the input. That is, the VL concepts and skills are distributed across different layers. Hence, for the clustering, we choose five layers spanning from the initial to top layers of the model to cover a wide range of concepts and skills and use the concatenation of their output as the feature vector of each data point.

We cluster VIT training data points using their feature vector from multiple layers of a small LVLM, called a reference model. We extract the features right after the MSA of the $l$-th layer (Eq.~\ref{eq:tokens_processing}) and process them into unit-length vectors:
\begin{align}
    \hspace{-2mm}
    \begin{split}
    \bm{u}^{v}_{l}&=\texttt{L2-Normalize}(\texttt{MeanPool}(\texttt{tanh}(\bm{z}^{v}_{l}))), \\
    \bm{u}^{t}_{l}&=\texttt{L2-Normalize}(\texttt{MeanPool}(\texttt{tanh}(\bm{z}^{t}_{l}))),
    \label{eq:unit_vectorize}
    \end{split}
    \hspace{-2mm}
\end{align}
where the mean-pooling is performed across the number of visual and text tokens, respectively. The hyperbolic tangent function, $\texttt{tanh}$, is necessary to reduce the impact of a few extreme activations, which are described by~\citet{Sun2024massiveact}. Without this step, these large values would dominate the feature vector and skew the clustering.
After that, we concatenate features from the small LVLM's layers:
\begin{equation}
    \bm{u}^{m}=[\bm{u}^{v}_{l_{1}},\;\bm{u}^{t}_{l_{1}},\;\ldots,\;\bm{u}^{v}_{l_{M}},\;\bm{u}^{t}_{l_{M}}]\;/\;\sqrt{2M},
    \label{eq:multi_neuron_act}
\end{equation}
where $M$ denotes the number of layers where we extract the features, and the subscripts $l_{1}, \ldots l_{M}$ are the layer indices. The resultant $\bm{u}^{m} \in \mathbb{R}^{2M*D}$ is the final multimodal feature of the data point. 

Then, we perform spherical k-means clustering on $\bm{u}^{m}$, yielding $K$ clusters. To ensure the purity of clusters, we set $K$ to a large number, such as $10,000$. Despite its simplicity, the k-means procedure runs in $O(NK)$ time for $N$ data points, which is advantageous when both $N$ and $K$ are large. Other clustering techniques such as spectral clustering or affinity propagation are much more expensive. Qualitative analysis indicates the clusters coincide with concept-skill compositions. We provide visualization of the clusters in~\Cref{appendix:cluster_visualize}.

\subsection{Measuring Cluster Transferability \label{sec:subsec:correlation_discover}}
\begin{figure}[t]
    \centering
    \small
    \includegraphics[width=\linewidth]{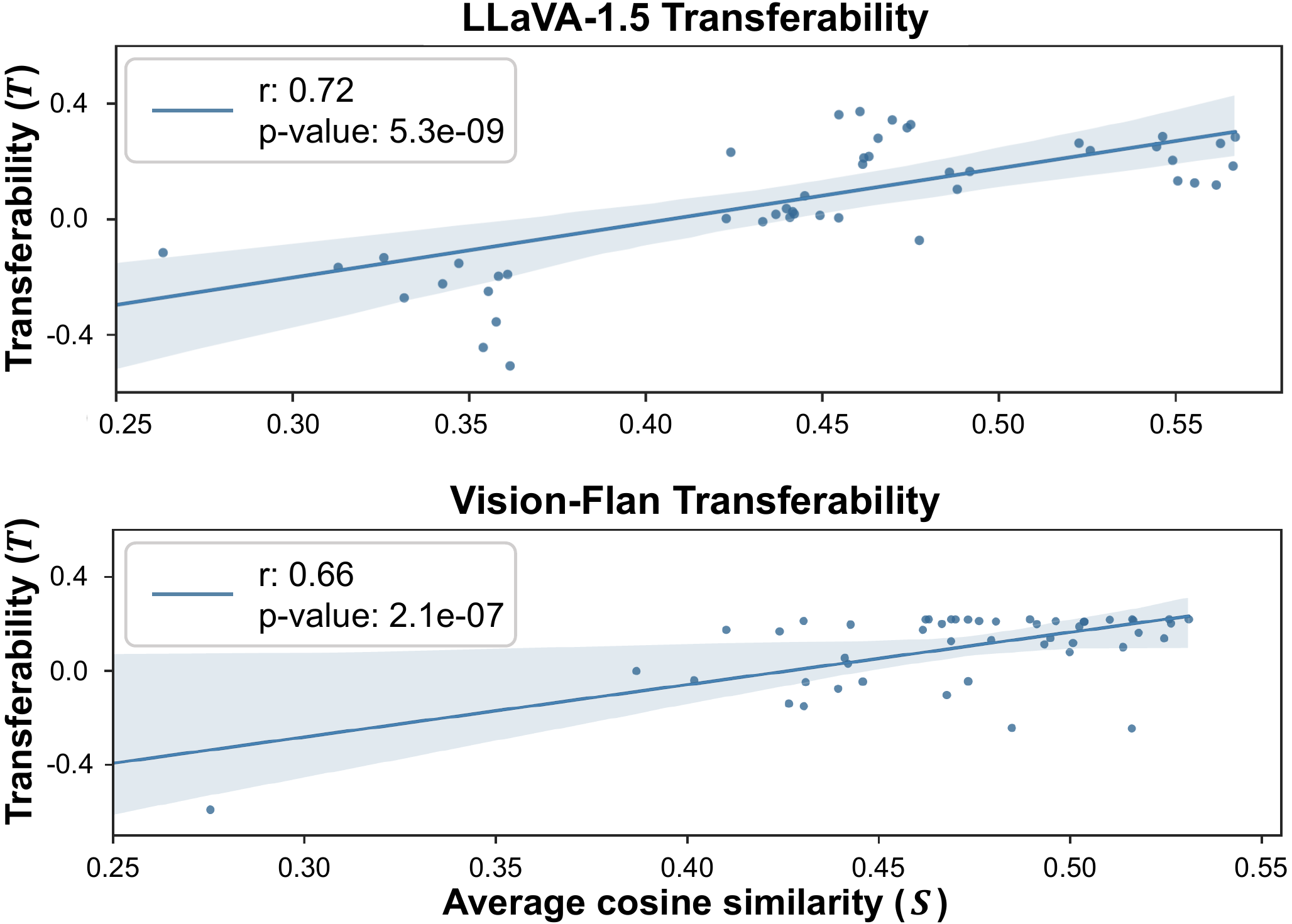}
    \par
    \caption{Correlation between cluster centroid similarity and transferability. We examine the correlations in the LLaVA 1.5~\citep{Liu2023llava15} and Vision-Flan~\citep{Xu2024visionflan} datasets, with each point representing a source cluster. We report the Pearson correlation coefficient ($r$) and p-value.}
    \label{fig:sim_gain_correlation}
\end{figure}

Empirical evidence shows that datasets differ in their ability to generalize to other datasets~\citep{Zamir_2018_CVPR, achille2020information}. We hypothesize that (1) data clusters also have varying levels of transferability and (2) clusters close together in feature space transfer well to each other. If (1) is true, it would be beneficial to select data from highly transferable clusters. If (2) is true, we can use distance among clusters as a proxy for transferability.

We design an experiment to verify the hypotheses. Following \citet{NEURIPS2023_70b8505a}, to measure transferability from cluster $C_i$ to cluster $C_j$, we run two training sessions. In the first, we finetune an LVLM on the same number of samples, $N_c$, drawn from $C_i$ and $C_j$ respectively. In the second, we finetune on $N_c$ samples from $C_j$ only. After finetuning, both models are tested on unseen samples from $C_j$, yielding test losses $L_{i,j\rightarrow j}$ and  $L_{j\rightarrow j}$. The difference $L_{j\rightarrow j}-L_{i,j\rightarrow j}$ can be seen as the degree by which $C_i$ facilitates the learning of $C_j$. We aggregate over target clusters to compute the transferability of the source cluster $C_i$:
\begin{equation}
T_i = \frac{1}{K_{\text{tgt}}} \sum_{j=1}^{K_{\text{tgt}}} (L_{j\rightarrow j}-L_{i,j\rightarrow j}),
\label{eq:eval_gain}
\end{equation}
where $K_{\text{tgt}}$ is the number of target clusters. Then, we compute the cosine similarity of the source cluster with the target clusters and average:
\begin{equation}
S_i = \frac{1}{K_{\text{tgt}}} \sum_{j=1}^{K_{\text{tgt}}} \text{cos}(\bm{e}_i, \bm{e}_j),
\label{eq:cosine_sim}
\end{equation}
where $\bm{e}_{i}$ is the cluster centroid of cluster $C_i$.

We compute the correlation between transferability $T_i$ and average cosine similarity $S_i$ over all possible pairings between 50 random source clusters and 50 random target clusters, and plot the results in~\Cref{fig:sim_gain_correlation}. We find that (1) clusters differ significantly in transfer power, and (2) $S_i$ and $T_i$ have a strong positive correlation (0.66-0.72), indicating that the cosine similarity among clusters can serve as an effective and inexpensive proxy for transferability. For $K$ clusters, the time complexity of all cosine similarities is $O(K^2)$. Further studies of transferability are available in~\Cref{appendix:in_depth_concept_skill_analysis}.

\subsection{Data Selection Criteria \label{sec:subsec:our_approach}}
In addition to transferability $T_i$ and its proxy $S_i$, we consider the density of a cluster during the sampling process, as selecting too many data points from a dense cluster that contains many similar samples would create redundancy. Hence, we introduce a density measure $D_i$:
\begin{equation}
D_i = \frac{1}{|C_i|(|C_i| - 1)} \sum_{p, q \in C_i, p\neq q} d(p, q),
\label{eq:density_cal}
\end{equation}
where $p$ and $q$ are two distinct data points from cluster $C_i$, and $d(p, q)=\exp(-\lVert\bm{u}^{m}_{p}-\bm{u}^{m}_{q}\rVert^{2})$ is the Gaussian kernel function with $\bm{u}^{m}_{p}$ and $\bm{u}^{m}_{q}$ being the multimodal neuron activations (Eq.~\ref{eq:multi_neuron_act}) of data points $p$ and $q$, respectively. The small $D_i$ value indicates that the cluster $C_i$ is highly diverse. 

In order to create a coreset of $N_{\text{core}}$ samples, we select from cluster $C_i$ exactly $N_{\text{core}} P_i$ samples. Here, $P_i \propto \exp(S_i / ( \tau D_i))$ is a categorical distribution and $\tau$ is a temperature hyperparameter. This approach enables us to select more samples from more transferable and less dense clusters to enhance training efficacy, while still selecting a few samples from other clusters to ensure diverse concept-skill compositions in the coreset. 

From cluster $C_i$, we aim to select $N_{\text{core}} P_i$ samples that are representative of the original data distribution of $C_i$. We compute the distance between the original cluster $C_i$ and the set of sampled data points $C^{\prime}_{i}$ as $\text{MMD}^2$, the squared maximum mean discrepancy, which is defined as:
\begin{align}
    \text{MMD}^{2}\!=&A(C_{i},C_{i})\!+\!A(C^{\prime}_{i},{C}^{\prime}_{i})\!-\!2A(C_{i},C^{\prime}_{i}), \notag \\
    A(C_{i},C_{j}&)\!=\!\frac{1}{|C_i||C_j|} \sum_{p \in C_i, q \in C_j} d(p, q). \label{eq:mmd_cal}
\end{align}
We iteratively add samples from the cluster $C_i$ to the sampled cluster $C^{\prime}_{i}$ that minimizes $\text{MMD}^2$ using greedy search~\citep{Kim2016mmdcritic}. In the end, we combine all the selected samples from all the clusters to compose the final VIT coreset. The complete data selection algorithm is shown in~\Cref{appendix:algorithm}.

\section{Experiments}
\begin{table*}[t]
    \tiny
    \caption{Comparison of coreset selection techniques on the LLaVA-1.5 dataset. We finetune the models using coresets with a 20\% sampling ratio and estimate performance on various multimodal evaluation benchmarks. The best and the second best results are in \textbf{bold} and \underline{underlined}, respectively.}
    \centering
    \resizebox{\textwidth}{!}{
        \renewcommand{\arraystretch}{1.25}
        \renewcommand{\tabcolsep}{5.0pt}
        \begin{tabular}{l c c c c c c c cc c a}
             \toprule
             {\textbf{Method}} & {\textbf{VQAv2}} & {\textbf{GQA}} & {\textbf{VizWiz}} & {\textbf{SQA-I}} & {\textbf{TextVQA}} & {\textbf{POPE}} & {\textbf{MME}} & \multicolumn{2}{c}{\textbf{MMBench}} & {\textbf{LLaVA-}} & {\textbf{\relp (\%)}}\\
             & & & & & & & & {\textbf{en}} & {\textbf{cn}} & {\textbf{Bench}} & \\
             \midrule
             Full-Finetune &
             {\scriptsize 79.1} & {\scriptsize 63.0} & {\scriptsize 47.8} & {\scriptsize 68.4} & {\scriptsize 58.2} & {\scriptsize 86.4} & {\scriptsize 1476.9} & {\scriptsize 66.1} & {\scriptsize 58.9} & {\scriptsize 67.9} & {\scriptsize 100}\\
             \cmidrule{0-11}
              Random &
             {\scriptsize 75.7} & {\scriptsize 58.9} & {\scriptsize 44.3} & {\scriptsize 68.5} & {\scriptsize 55.3} & {\scriptsize 84.7} & {\scriptsize \underline{1483.0}} & {\scriptsize 62.2} & {\scriptsize \underline{54.8}} & {\scriptsize 65.0} & {\scriptsize \underline{95.8}}\\
                CLIP-Score &
             {\scriptsize 73.4} & {\scriptsize 51.4} & {\scriptsize 43.0} & {\scriptsize 65.0} & {\scriptsize 54.7} & {\scriptsize 85.3} & {\scriptsize 1331.6} & {\scriptsize 55.2} & {\scriptsize 52.0} & {\scriptsize 66.2} & {\scriptsize 91.2}\\
              EL2N &
             {\scriptsize \underline{76.2}} & {\scriptsize 58.7} & {\scriptsize 43.7} & {\scriptsize 65.5} & {\scriptsize 53.0} & {\scriptsize 84.3} & {\scriptsize 1439.5} & {\scriptsize 53.2} & {\scriptsize 47.4} & {\scriptsize 64.9} & {\scriptsize 92.0}\\
              Perplexity &
             {\scriptsize 75.8} & {\scriptsize 57.0} & {\scriptsize \underline{47.8}} & {\scriptsize 65.1} & {\scriptsize 52.8} & {\scriptsize 82.6} & {\scriptsize 1341.4} & {\scriptsize 52.0} & {\scriptsize 45.8} & {\scriptsize \underline{68.3}} & {\scriptsize 91.6}\\
                SemDeDup &
             {\scriptsize 74.2} & {\scriptsize 54.5} & {\scriptsize 46.9} & {\scriptsize 65.8} & {\scriptsize \underline{55.5}} & {\scriptsize 84.7} & {\scriptsize 1376.9} & {\scriptsize 52.2} & {\scriptsize 48.5} & {\scriptsize \textbf{70.0}} & {\scriptsize 92.6}\\
                D2-Pruning &
             {\scriptsize 73.0} & {\scriptsize 58.4} & {\scriptsize 41.9} & {\scriptsize \textbf{69.3}} & {\scriptsize 51.8} & {\scriptsize \underline{85.7}} & {\scriptsize 1391.2} & {\scriptsize \textbf{65.7}} & {\scriptsize \textbf{57.6}} & {\scriptsize 63.9} & {\scriptsize 94.8}\\
                Self-Sup &
             {\scriptsize 74.9} & {\scriptsize \underline{59.5}} & {\scriptsize 46.0} & {\scriptsize 67.8} & {\scriptsize 49.3} & {\scriptsize 83.5} & {\scriptsize 1335.9} & {\scriptsize 61.4} & {\scriptsize 53.8} & {\scriptsize 63.3} & {\scriptsize 93.4}\\
                Self-Filter &
             {\scriptsize 73.7} & {\scriptsize 58.3} & {\scriptsize \textbf{53.2}} & {\scriptsize 61.4} & {\scriptsize 52.9} & {\scriptsize 83.8} & {\scriptsize 1306.2} & {\scriptsize 48.8} & {\scriptsize 45.3} & {\scriptsize 64.9} & {\scriptsize 90.9}\\
                \cellcolor{gg}\ours (Ours) &
             \cellcolor{gg}{\scriptsize \textbf{76.5}} & \cellcolor{gg}{\scriptsize \textbf{59.8}} & \cellcolor{gg}{\scriptsize 46.8} & \cellcolor{gg}{\scriptsize \underline{69.2}} & \cellcolor{gg}{\scriptsize \textbf{55.6}} & \cellcolor{gg}{\scriptsize \textbf{86.1}} & \cellcolor{gg}{\scriptsize \textbf{1495.6}} & \cellcolor{gg}{\scriptsize \underline{63.1}} & \cellcolor{gg}{\scriptsize 54.5} & \cellcolor{gg}{\scriptsize 67.3} & \cellcolor{gg}{\scriptsize \textbf{97.4}}\\
             \bottomrule
        \end{tabular}
    }
    \label{tab:llava_eval}
\end{table*}

\subsection{Setup\label{sec:subsec:exp_setup}}
\paragraph{Visual Instruction Tuning Datasets}
We conduct coreset selection on two distinct VIT datasets: LLaVA-1.5~\citep{Liu2023llava15} and Vision-Flan~\citep{Xu2024visionflan}. The LLaVA-1.5 dataset contains 665k VIT data from 12 different VL tasks. The Vision-Flan dataset comprises 191 VL tasks, each with approximately 1k expert-annotated VIT data points, totaling 186k samples.

\paragraph{Models for Training and Data Selection}
For the target LVLMs, we use the pre-trained LLaVA-1.5 model~\citep{Liu2023llava15} with a default size of 7B parameters unless otherwise specified. In all experiments, we train the models using LoRA~\citep{Hu2022lora} for one epoch, following the official finetuning hyperparameters specified in LLaVA-1.5. As a reference model, we use the TinyLLaVA-2B~\citep{Zhou2024tinyllava}, a small LVLM finetuned on the target VIT dataset, for efficient coreset selection for all methods unless otherwise specified. All experiments are conducted using 4 V100 GPUs.

\paragraph{Evaluation Benchmark} To assess the generalization of finetuned LVLMs across diverse visual instructions, we evaluate the models on several widely adopted zero-shot multimodal evaluation benchmarks, including  
1) visual question answering: VQAv2~\citep{vqav2}, GQA~\citep{gqa}, VizWiz~\citep{vizwiz}; 2) knowledge-grounded QA: ScienceQA~\citep{sqa}; 3) Optical Character Recognition (OCR): TextVQA~\citep{textvqa}; 4) hallucination: POPE~\citep{pope}; 5) multiple-choice: MME~\citep{mme}, MMBench~\citep{mmbench}; 6) free-form generation: LLaVA-Bench~\citep{Liu2023llava}, MM-Vet~\citep{mmvet}. In all experiments, we follow the protocols outlined in LLaVA-1.5 and Vision-Flan to select evaluation benchmarks. Further explanations of these benchmarks are provided in~\Cref{appendix:setups}.

Since each evaluation benchmark has a different scale, we compute average relative performance, denoted as \relp, across benchmarks to assess the level of generalization. Each relative performance is derived from the formula: (model performance / full-finetuned performance) $\times$ 100\%.

\paragraph{Baselines}
We compare our method with several coreset selection techniques: CLIP-Score, EL2N~\citep{Paul2021el2n}, Perplexity~\citep{Marion2023llmpre}, SemDeDup~\citep{Abbas2023semdedup}, D2-Pruning~\citep{Maharana2023d2prune}, Self-Sup~\citep{Sorscher2022scalinglaw}. We also compare with a recent VIT coreset selection method, Self-Filter~\citep{Chen2024selffilter}. We additionally report the results of \textit{Random}, the model finetuned with the coreset collected by random sampling, and \textit{Full-Finetune}, the model finetuned with the full VIT dataset. The details of the baseline methods are provided in~\Cref{appendix:setups}.

\subsection{Results\label{sec:subsec:exp_results} and Discussion}
\paragraph{\ours\ surpasses baselines on LLaVA-1.5.}
\Cref{tab:llava_eval} presents model performance when we limit the coreset to 20\% of the size of the LLaVA-1.5 VIT dataset. \ours is either the best or a close second on 7 out of 10 benchmarks, including VQAv2, GQA, SQA-I, TextVQA, POPE, MME, and MMBench-en. On average, \ours outperforms the best baseline by 1.6 percent points (pp) in relative performance.

Interestingly, all baselines perform worse than the random sampling on average relative performance, suggesting that they may be susceptible to the selection bias, which is discussed in the introduction and illustrated in~\Cref{fig:score_distribution}. In contrast, \ours considers the diversity of VL concept-skill compositions, demonstrating high generalization across a broad range of visual instructions. We further analyze the selection bias of the baselines and effectiveness of \ours in~\Cref{appendix:diversity_analysis}.

In \Cref{fig:llava_selection_ratio}, we show the performance comparison across different coreset sizes as proportions of the original LLaVA-1.5 dataset. \ours consistently outperforms other baselines across various sampling ratios, underscoring the effectiveness of our approach. \ours also performs well on LLaVA-1.5-13B, as shown in~\Cref{appendix:subsec:transfer_larger}.

\begin{figure}[t]
    \centering
     \includegraphics[width=\linewidth]{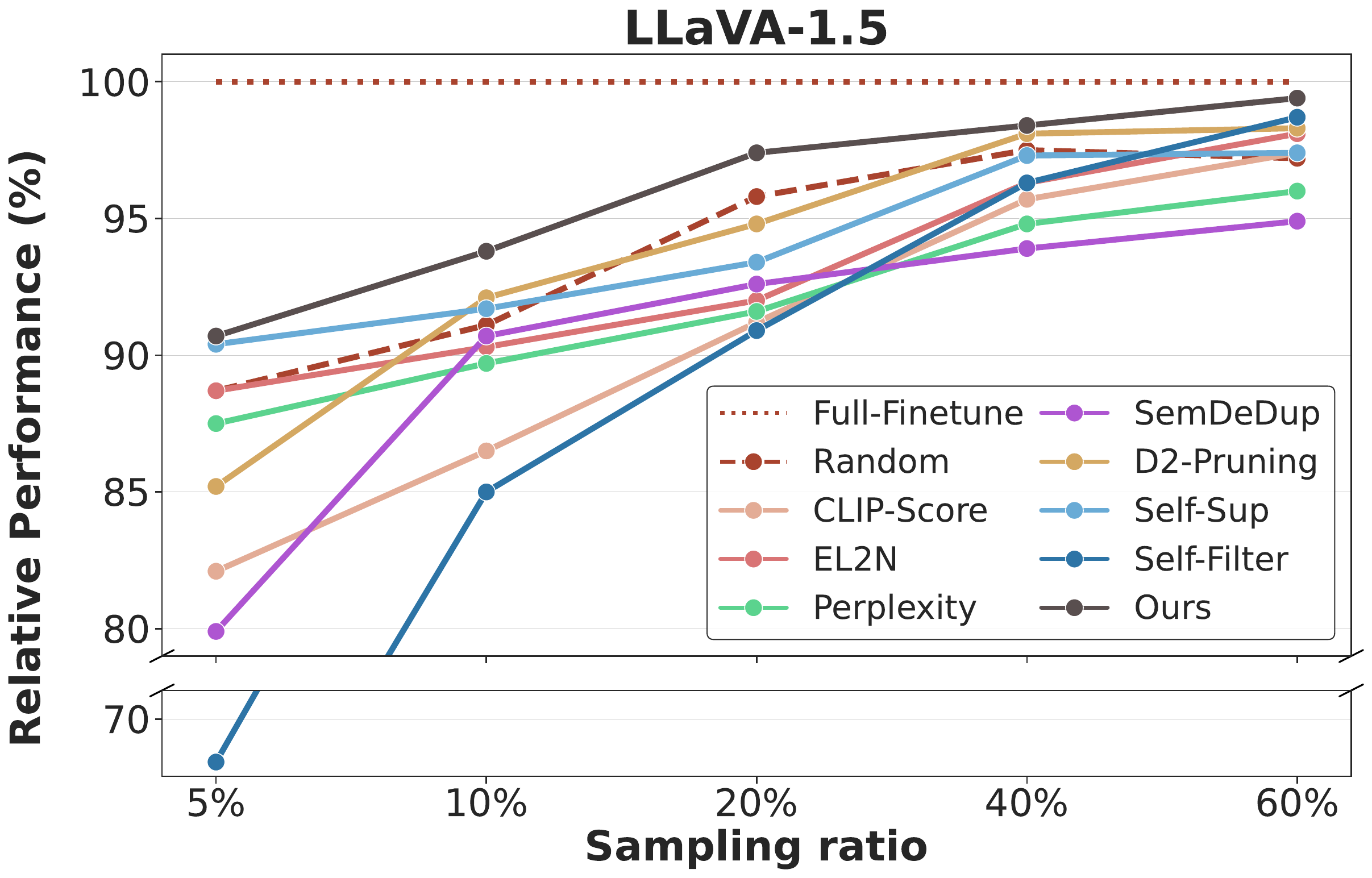}
     \par
    \caption{Average relative performances of all coreset selection techniques at different sampling ratios for the LLaVA-1.5 dataset.}
    \label{fig:llava_selection_ratio}
\end{figure}
\begin{figure}[t]
    \centering
    \includegraphics[width=\linewidth]{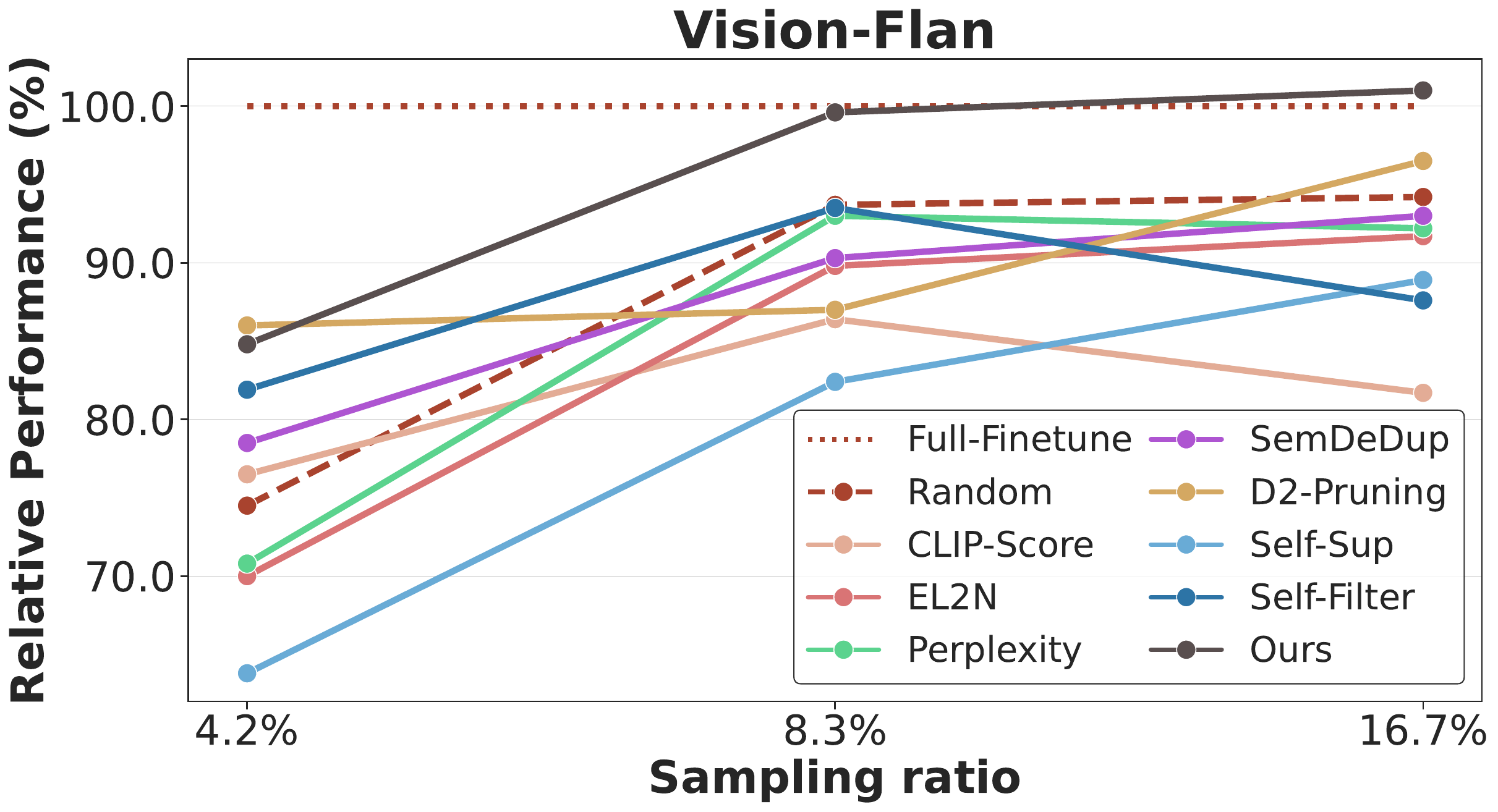}
    \par
    \caption{Average relative performances of all coreset selection techniques at different sampling ratios for the Vision-Flan dataset.}
    \label{fig:vision_flan_selection_ratio}
\end{figure}

\begin{table}[t]
    \tiny
    \caption{Comparison of coreset selection techniques on the Vision-Flan dataset. We finetune the models using coresets with a 16.7\% sampling ratio and estimate performance on various multimodal evaluation benchmarks. The best and the second best results are in \textbf{bold} and \underline{underlined}, respectively.}
    \centering
    \resizebox{\linewidth}{!}{
        \renewcommand{\arraystretch}{1.3}
        \renewcommand{\tabcolsep}{1.0pt}
        \begin{tabular}{l c c c c c a}
             \toprule
             {\textbf{Method}} & {\textbf{MMBench-en}} & {\textbf{MME}} & {\textbf{MM-Vet}} & {\textbf{POPE}} & {\textbf{SQA-I}} & {\textbf{\relp (\%)}}\\
             \midrule
             Full-Finetune &
             {\scriptsize 53.4} & {\scriptsize 1287.5} & {\scriptsize 25.6} & {\scriptsize 84.2} & {\scriptsize 61.3} & {\scriptsize 100}\\
             \cmidrule{0-6}
              Random &
             {\scriptsize 45.2} & {\scriptsize 1122.3} & {\scriptsize 26.1} & {\scriptsize 82.5} & {\scriptsize 60.9} & {\scriptsize 94.2}\\
                CLIP-Score &
             {\scriptsize 34.3} & {\scriptsize 687.6} & {\scriptsize 26.6} & {\scriptsize 72.6} & {\scriptsize 61.8} & {\scriptsize 81.7}\\
              EL2N &
             {\scriptsize 45.3} & {\scriptsize 1082.9} & {\scriptsize 23.9} & {\scriptsize 82.1} & {\scriptsize 60.6}  & {\scriptsize 91.7}\\
              Perplexity &
             {\scriptsize 39.3} & {\scriptsize \underline{1160.9}} & {\scriptsize 26.1} & {\scriptsize \underline{83.1}} & {\scriptsize 59.2} & {\scriptsize 92.2}\\
                SemDeDup &
             {\scriptsize 42.1} & {\scriptsize 1146.5} & {\scriptsize \underline{27.2}} & {\scriptsize 82.7} & {\scriptsize 56.8} & {\scriptsize 93.0}\\
                D2-Pruning &
             {\scriptsize \underline{49.1}} & {\scriptsize 1052.4} & {\scriptsize 27.0} & {\scriptsize 82.5} & {\scriptsize \textbf{64.7}} & {\scriptsize \underline{96.5}}\\
                Self-Sup &
             {\scriptsize 42.9} & {\scriptsize 1012.2} & {\scriptsize 23.5} & {\scriptsize 80.8} & {\scriptsize 60.0} & {\scriptsize 88.9}\\
                Self-Filter &
             {\scriptsize 28.6} & {\scriptsize 923.6} & {\scriptsize \textbf{30.0}} & {\scriptsize \textbf{83.3}} & {\scriptsize 59.3} & {\scriptsize 87.6}\\
                \cellcolor{gg}\ours (Ours) &
             \cellcolor{gg}{\scriptsize \textbf{56.7}} & \cellcolor{gg}{\scriptsize \textbf{1222.2}} & \cellcolor{gg}{\scriptsize 26.2} & \cellcolor{gg}{\scriptsize 81.9} & \cellcolor{gg}{\scriptsize \underline{63.8}} & \cellcolor{gg}{\scriptsize \textbf{101.0}}\\
             \bottomrule
        \end{tabular}
    }
    \label{tab:vision_flan_eval}
\end{table}
\paragraph{One Sixth of Vision-Flan selected by \ours\ outperforms full dataset.} We further evaluate the coreset selection techniques on the Vision-Flan VIT dataset~\citep{Xu2024visionflan} and show the results in \Cref{tab:vision_flan_eval}. \ours exceeds the performance of the model finetuned on the whole Vision-Flan data by 1.0 pp and the performance of the best baseline by 4.5 pp, using a selected subset 16.7\% (1/6) of its size. Further, as illustrated in~\Cref{fig:vision_flan_selection_ratio}, \ours maintains consistently high performance across several sampling rates.

We note that Vision-Flan, with its 191 VL tasks, is much more diverse than the LLaVA-1.5 dataset of 12 tasks. The stronger performance of \ours on the Vision-Flan suggests that \ours algorithm is well adapted to the use case of visual instruction tuning, which is increasingly performed on larger and more diverse sets of tasks. 

Another curious phenomenon is that several baselines, including CLIP-Score, Perplexity, and Self-Filter, experience performance declines as the sampling ratio increases in~\Cref{fig:vision_flan_selection_ratio}. A similar trend is observed in the random baseline in~\Cref{fig:llava_selection_ratio}. This underscores the importance of deliberate coreset selection, as merely increasing the dataset size does not guarantee improved LVLM capabilities.
\begin{figure}[t]
    \centering
    \includegraphics[width=\linewidth]{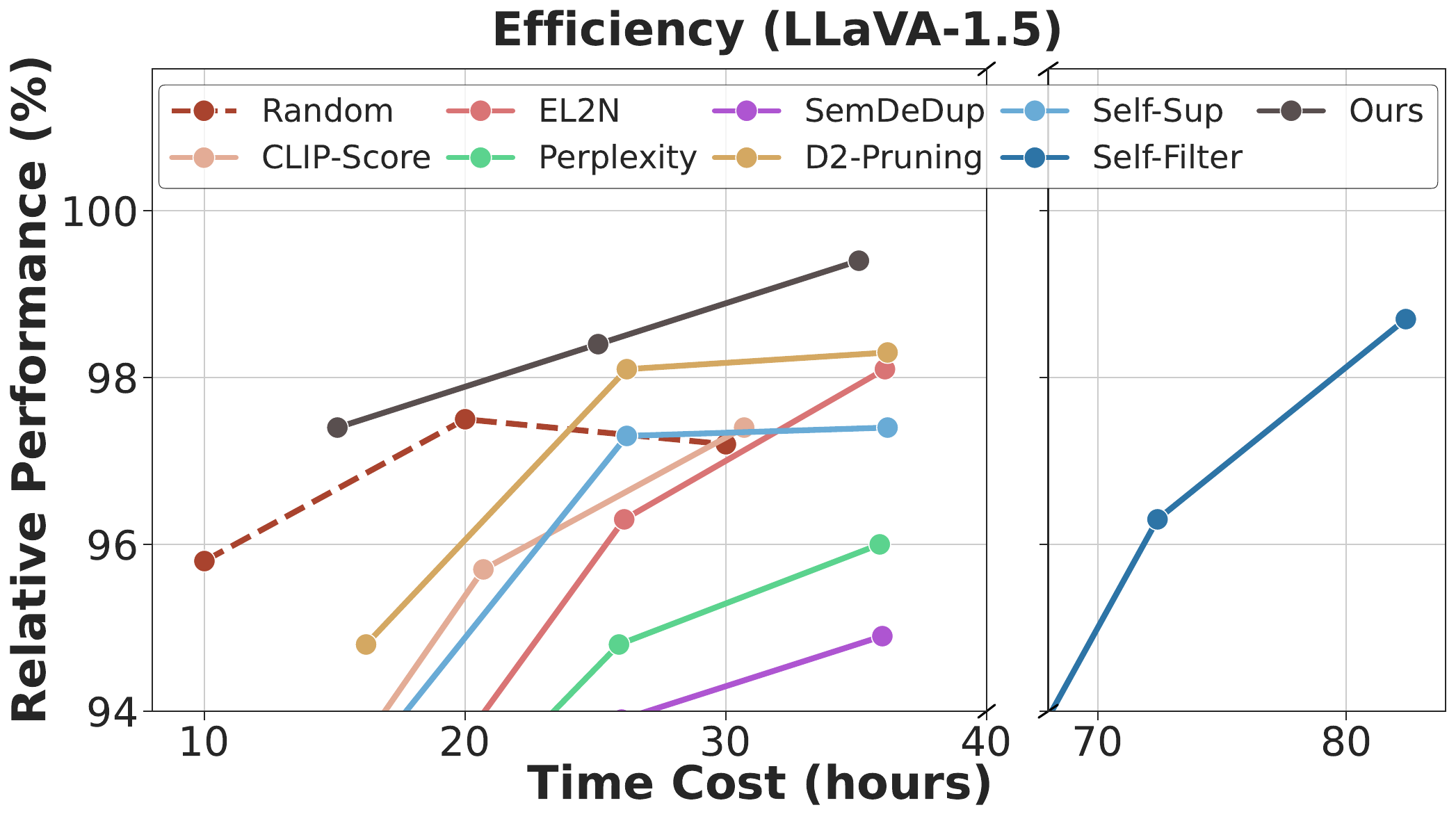}
    \par
    \caption{Comparison of coreset selection techniques on average relative performance and wall-clock time cost. The wall-clock time cost includes both the data selection and finetuning of the target LVLM. The time cost is measured in hours of running time on a computing node with 4$\times$ V100 GPUs.}
    \label{fig:efficiency}
\end{figure}

\begin{table*}[t]
    \centering
    \caption{Ablation studies of \ours. \textbf{(a)} Effect of different reference models. The time cost includes both the data selection and finetuning of the target LVLM and is measured in hours of running time on a computing node with 4$\times$ V100 GPUs. \textbf{(b)} Ablation on data selection criteria of our approach, transferability ($S$) and density ($D$). \textbf{(c)} The performance of different intra-cluster sampling strategies across various coreset sizes.}
    \vspace{0.025in}
    \label{tab:ablation_merge}
    \begin{minipage}{0.27\textwidth}
        \centering
        \small{\textbf{(a) Reference Model}}\\
        \vspace{0.1in}
        \resizebox{\textwidth}{!}{
    \renewcommand{\arraystretch}{1.35}
    \renewcommand{\tabcolsep}{2.0pt}
    \begin{tabular}{l c c}
         \toprule
         {\textbf{Model}} & {\textbf{Time}} & {\textbf{\relp }}\\
         {\textbf{(\# params)}} & {\textbf{(hours)}} & {\textbf(\%)} \\
         \midrule
         CLIP (0.4B) & \textbf{10.9} & 94.2\\
         TinyLLaVA (0.9B) & 12.2 & 96.3\\
         TinyLLaVA (2B) & 15.3 & \textbf{97.4}\\
         LLaVA-1.5 (7B) & 20.7 & 97.1\\
         \bottomrule
    \end{tabular}
}
\label{tab:ablation_reference_model}

    \end{minipage}
    \begin{minipage}{0.29\textwidth}
        \centering
        \small{\textbf{(b) Key Components}}\\
        \vspace{0.1in}
        \resizebox{\linewidth}{!}{%
\renewcommand{\arraystretch}{1.27}
\renewcommand{\tabcolsep}{2.5pt}
\begin{tabular}{lccc}
     \toprule
     \textbf{Method} & $\bm{S}$ & $\bm{D}$ & \textbf{\relp (\%)}\\ 
     \midrule
     Random & $-$ & $-$ & 95.8\\
     \midrule
     \multirow{4}{*}{\ours (Ours)} & $-$ & $-$ & 94.4\\
     & $\checkmark$ & $-$ & 95.9\\
     & $-$ & $\checkmark$ & 94.7\\
     & $\checkmark$ & $\checkmark$ & \textbf{97.4}\\
     \bottomrule
\end{tabular}
}

    \end{minipage}
    \begin{minipage}{0.42\textwidth}
        \centering
        \small{\textbf{(c) Intra-Cluster Sampling methods}}\\
        \vspace{0.1in}
        \resizebox{\linewidth}{!}{
\renewcommand{\arraystretch}{1.6}
\renewcommand{\tabcolsep}{1.5pt}
\begin{tabular}{l ccccc}
     \toprule
     {\textbf{Intra-Cluster Sampling}} & \multicolumn{5}{c}{\textbf{Sampling ratio}}\\
     & \textbf{5\%} & \textbf{10\%} & \textbf{20\%} & \textbf{40\%} & \textbf{60\%}\\
     \midrule
     Random-select & 90.1 & \textbf{94.3} & \textbf{97.5} & 97.7 & \underline{98.3} \\
     Nearest-to-centroid & \textbf{91.9} & \textbf{94.3} & 96.7 & \textbf{99.1} & \underline{98.4} \\
     Greedy-MMD$^{2}$-minimize & \underline{90.7} & 93.8 & \textbf{97.4} & \underline{98.4} & \textbf{99.4}\\
     \bottomrule
\end{tabular}
}

    \end{minipage}
    \vspace{0.05in}
\end{table*}
\paragraph{\ours\ provides wall-clock training time reduction and is Pareto superior.} In~\Cref{fig:efficiency}, we plot the wall-clock time cost of the entire pipeline of data selection and model finetuning versus the average relative performance (\relp) on the LLaVA-1.5 dataset. \ours achieves 97.4\%, 98.4\%, and 99.4\% relative performance with the wall-clock times of 15.1, 25.1, and 35.1 hours, respectively. In contrast, finetuning on all data takes 50 hours.

We observe that \ours provides Pareto superior solutions to all baselines. This is mainly due to the excellent time complexity of \ours, which is linear to the number of training data points. Moreover, our method discovers the transferability among clusters at a low computational cost. It requires only cosine similarity calculations, with a time complexity quadratic to the number of clusters. Hence, \ours provides a scalable data selection procedure.

\ours also utilizes neuron activations from intermediate layers of the small reference model rather than the final outputs, avoiding complete forward passes like other baselines. Additionally, \ours does not require training of additional networks that score data points, like Self-Filter. Neither does it require backward passes like the concurrent work TIVE~\citep{Liu2024tive}. The combination of all these factors leads to an efficient solution to coreset selection.

\subsection{Further Analysis and Ablation \label{sec:subsec:analysis}}
\paragraph{Alternative Reference Models} We analyze the effects of different reference models, which are the models used to extract features for clustering and cosine similarity. We compare four models, CLIP, TinyLLaVA-0.9B, TinyLLaVA-2B, and LLaVA-1.5-7B, and report the time cost of the entire coreset selection pipeline and average relative performance in~\Cref{tab:ablation_merge} \highlight{(a)}. We observe that CLIP performs the worst whereas TinyLLaVA-2B performs the best with reasonable time cost in data selection. However, the differences between TinyLLaVA-0.9B, TinyLLaVA-2B, and LLaVA-1.5-7B are small. We conclude that a well-trained small model can serve effectively as a reference model in coreset selection for a target LVLM. We also examine the robustness of \ours when the reference model is finetuned on a different VIT dataset, which is detailed in~\Cref{appendix:subsec:robust_reference}.

\paragraph{Ablation on Data Selection Criteria} To validate our coreset selection method, we conduct ablation studies on the two data selection criteria, transferability and density, as summarized in~\Cref{tab:ablation_merge} \highlight{(b)}. In the first ablation, without using either criterion, we simply select the same number of samples from each cluster. This results in inferior performance, which suggests that naive stratified sampling from the clusters is not sufficient, possibly due to the heterogeneous nature of the clusters. In the second ablation, number of samples from each cluster is proportional to the transferability of the cluster, leading to a 1.5 percentage point (pp) increase. The third ablation selects number of samples inversely proportional to density, yielding a modest enhancement of 0.3 pp. Finally, combining both transferability and density provides a sizeable increase of 3.0 pp, demonstrating that the two selection criteria are complementary to each other.

\paragraph{Intra-cluster Selection Criteria} \ours selects samples within a cluster by minimizing MMD$^2$. We examine the effects of two alternative techniques, random selection and selecting samples closest to the centroids. As shown in~\Cref{tab:ablation_merge} \highlight{(c)}, in small coresets, samples closest to the centroids, which are probably not outliers or hard samples, lead to high performance. In contrast, under high sampling ratios (i.e., large coresets), selecting diverse data using the MMD$^2$ metric leads to high performance. This is reminiscent of the finding of~\citet{Sorscher2022scalinglaw} that easy samples are beneficial when the sampling ratio is small, whereas hard samples are advantageous when the sampling ratio is large. Overall, \ours is robust to the choice of intra-cluster sampling, but adapting the intra-cluster sampling method to the sampling ratio can enhance the effectiveness of our approach.

\section{Conclusion}
In this paper, we introduce \ours, a cluster-level data selection technique for efficient visual instruction tuning of Large Vision-Language Models. We demonstrate that clustering based on internal activations from a small model can represent visual-linguistic concept-skill compositions shared among diverse tasks in visual instruction tuning datasets. Additionally, our empirical investigation validates a strong positive correlation between cosine similarity and transferability among clusters. Based on the transferability and density of clusters, \ours selects more samples from more transferable and less dense clusters to enhance training efficacy, while preserving the diversity of concept-skill compositions within the coreset to ensure better model generalization ability. Comprehensive experiments on the LLaVA-1.5 and Vision-Flan datasets demonstrate that our method outperforms baselines across several benchmarks with the lowest data selection cost, showcasing its effectiveness and efficiency. The success of \ours suggests redundancy in popular VIT datasets and underscores the importance of a thorough understanding of data in training LVLMs. 


\section*{Limitations}
\label{sec:limitation}
In our experiments, we observe that VL concept-skill compositions are shared across various VL tasks and identify VL concept-skill compositions that transfer well to others. However, after identifying these compositions and performing coreset selection, we finetune the target LVLMs by randomly selecting samples from the coreset. Recognizing the growing research attention on the importance of training order in LLM instruction tuning, we believe that considering the training order for LVLMs is crucial to enhance efficiency in visual instruction tuning. In future research, we aim to develop a curriculum learning algorithm that automatically determines the optimal training order based on the identified VL concept-skill compositions to further reduce the development cost of a new model.

Additionally, we assess whether the data with similar concept-skill compositions are concentrated well on the clusters through human inspection. Therefore, further investigation should be conducted to quantitatively evaluate the clustering of data with similar concept-skill compositions, which may enable accurate identification of VL concept-skill compositions and accurate quantification of their transferability.

\section*{Ethics Statement}
In this work, we use publicly available visual instruction tuning datasets for coreset selection to enable easy replication. However, some data in the datasets contain erroneous answers about the visual content or images that do not clearly connect with the provided answers. Finetuning Large Vision-Language Models (LVLMs) with such data may lead to the generation of erroneous interpretations of images or hallucinations. 
This may pose an ethical issue for LVLM deployment in the real world. However, current coreset selection techniques, including ours, do not address hallucination in their selection processes. This motivates further research in coreset selection to identify visual instruction tuning data that minimizes hallucinations, aiming to build more reliable and trustworthy LVLMs.

\section*{Acknowledgements} 
Jaewoo Lee and Sung Ju Hwang are supported by the National Research Foundation of Korea (NRF) grant funded by the Korea government (MSIT) (No. RS-2023-00256259) and a grant of the Korea Machine Learning Ledger Orchestration for Drug Discovery Project (K-MELLODDY), funded by the Ministry of Health \& Welfare and Ministry of Science and ICT, Republic of Korea (No. RS-2024-12345678). Boyang Li is supported by the Nanyang Associate Professorship and  Fellowship (NRF-NRFF13-2021-0006) of the National Research Foundation, Singapore. Any opinions, findings, conclusions, or recommendations expressed in this material are those of the authors and do not reflect the views of the funding agencies.

\bibliography{custom}

\clearpage
\appendix

\section{Details of Experimental Setups}
\label{appendix:setups}

\paragraph{Evaluation Benchmark} We provide in-depth explanations of the multimodal evaluation benchmarks used in our experiments. (1) VQAv2~\citep{vqav2} evaluates the ability to understand and reason about general visual content by answering open-ended questions based on images. (2) GQA~\citep{gqa} assesses compositional reasoning and understanding skills, requiring models to understand relationships and attributes of objects within images. (3) Vizwiz~\citep{vizwiz} is designed to evaluate the model's ability to cope with real-world visual impairments. (4) ScienceQA-Image (SQA-I)~\citep{sqa} tests the model's science-related reasoning and visual understanding of images. (5) TextVQA~\citep{textvqa} specifically targets text in images, assessing the Optical Character Recognition (OCR) ability of models. (6) POPE~\citep{pope} measures object hallucination in models. (7) MME~\citep{mme} contains binary choice questions designed to evaluate perception and cognition abilities through 14 subtasks. (8) MMBench~\citep{mmbench} evaluates various abilities of models, covering object detection, text recognition, relation reasoning, etc., using tests conducted in English (en) or Chinese (cn). (9) LLaVA-Bench~\citep{Liu2023llava15} is specifically designed for evaluating models on visual instruction-following and chat ability. (10) MM-Vet~\citep{mmvet} measures VL capabilities, including recognition, OCR, knowledge, language generation, spatial awareness, and math.

\paragraph{Baselines} In this section, we provide a more detailed explanation of the baselines. The hyperparameters for baselines in our experiments are summarized in~\Cref{tab:baseline_hyp}.

\begin{itemize}[itemsep=2mm, parsep=1pt, leftmargin=*]
\item \textbf{CLIP-Score} utilizes the CLIP~\citep{Radford2021clip} model to assess the alignment between images and their instructions. For our study, we select VIT data with the highest CLIP scores.

\begin{table}[t]
\centering
\caption{Hyperparameter configurations.}
\vspace{-0.1in}
\resizebox{\linewidth}{!}{
    \renewcommand{\arraystretch}{1.25}
    \renewcommand{\tabcolsep}{5pt}
    \begin{tabular}{lll}
    \toprule
    Method  & LLaVA-1.5 & Vision-Flan \\ 
    \midrule
    CLIP-Score & high score selected & high score selected \\
    EL2N & medium score selected & medium score selected \\
    Perplexity & medium score selected & medium score selected \\
    SemDeDup & $\bm{K}$\;:\;10,000 & $\bm{K}$\;:\;5,000 \\
    D2-Pruning & $\bm{k}$\;:\;5,\;$\gamma_r$\;:\;0.4,\;$\gamma_f$\;:\;1.0 & $\bm{k}$\;:\;5,\;$\gamma_r$\;:\;0.4,\;$\gamma_f$\;:\;1.0 \\
    Self-Sup & $\bm{K}$\;:\;10,000 & $\bm{K}$\;:\;5,000 \\
    Self-Filter & $\bm{k}$\;:\;10,\;$\gamma$\;:\;1 & $\bm{k}$\;:\;10,\;$\gamma$\;:\;1 \\
    \ours (Ours) & $\bm{K}$\;:\;10,000,\;$\tau$\;:\;0.1 & $\bm{K}$\;:\;5,000,\;$\tau$\;:\;0.1 \\
    \bottomrule
    \end{tabular}
    }
\vspace{-0.125in}
\label{tab:baseline_hyp}
\end{table}
\item \textbf{EL2N}~\citep{Paul2021el2n} estimates sample quality using the Error L2-Norm score, defined as $\mathbb{E}[||p(x)-y||_2]$. Here, $p(\cdot)$ represents the reference model, $x$ is the input, and $y$ is the ground-truth label. This metric calculates the average L2 distance between the model's predictions and the ground-truth labels for text tokens.

\item \textbf{Perplexity}~\citep{Marion2023llmpre} measures the average negative log-likelihood of the next token prediction, defined as $\exp(-\mathbb{E}[\log p(x)])$. This metric assesses the uncertainty in the model's predictions. For both EL2N and Perplexity, we select data from the middle score distribution, as this range has been shown to perform best in prior research~\citep{Marion2023llmpre}.

\item \textbf{SemDeDup}~\citep{Abbas2023semdedup} removes semantically duplicated data by clustering the output embeddings of the last token from the reference model's final layer. This helps in reducing redundancy in the selected coreset.

\item \textbf{D2-Pruning}~\citep{Maharana2023d2prune} represents the dataset as a graph where nodes represent sample difficulty and edges represent distances between samples. It actively uses the graph to preserve diversity in the coreset. We use the AUM~\citep{Pleiss2020aum} score to indicate difficulty, defined as $p_y(x)-\underset{i\neq y}{\max}\;p_i(x)$, where $p_y(x)$ is the prediction value for the ground-truth label, and $\underset{i\neq y}{\max}\;p_i(x)$ is the highest prediction value for any non-ground-truth label. For the distances between samples, we calculate the L2 distance between averaged output embeddings from the last layer tokens of the reference model.

\item \textbf{Self-Sup}~\citep{Sorscher2022scalinglaw} clusters the data using the averaged output embeddings from the last layer tokens of the reference model. It scores data based on their distance to cluster centroids, selecting those the most likely to be prototypical.

\item \textbf{Self-Filter}~\citep{Chen2024selffilter} is a recent VIT coreset selection method that was originally applied to the LLaVA-158k VIT dataset~\citep{Liu2023llava}, which consists of only three VL tasks. It finetunes the score-net along with the target LVLM on the full dataset to serve as a reference model for scoring and filtering VIT data. We use the version that additionally incorporates both CLIP scores and CLIP features since it ensures enhanced performance and efficiency.

\begin{figure*}[t]
    \centering
    \includegraphics[width=\linewidth]{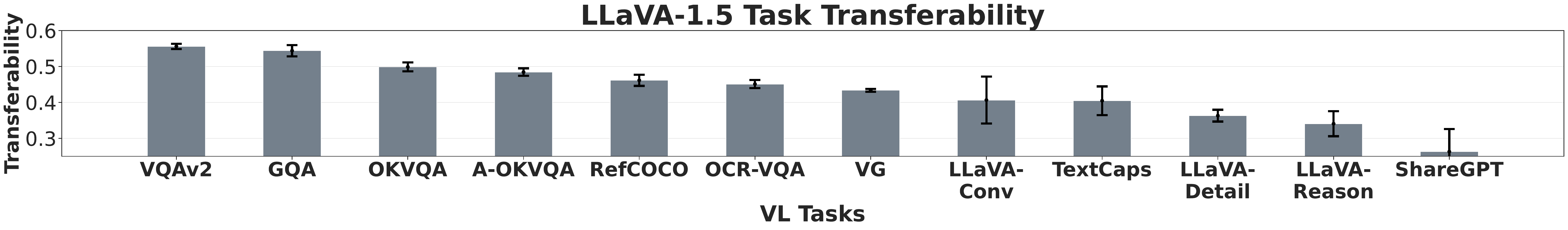}
    \par
    \caption{Task-wise transferability. We group the VIT data based on task names and then report the average cluster transferability of each group.}
    \label{fig:avg_transfer}
\end{figure*}

\end{itemize}

\section{Visualizing LVLM Skills with Relevancy Maps\label{appendix:relevancy_map}}
In our method, we extract neuron activations from various layers (Eq.~\ref{eq:unit_vectorize}) to represent the concepts and skills of each VIT data. In this approach, we hypothesize that distinct layers represent distinct concepts and skills of the LVLM. To support this assumption, we compute relevancy maps~\citep{Chefer2021relevancyscore} following the approach outlined in~\citet{Stan2024lvlminterpret}. The relevancy maps help us understand the model's final output by highlighting the most contributing parts of the input for each layer. Given the target output token $\bm{y_t}$ and the attention map $\bm{A}_{l}\!\in\!\mathbb{R}^{h\!\times\!(N_v+N_l)\times(N_v+N_l)}$ of the $\bm{l}$-th layer, where $h$ is the head dimension of the attention, the relevancy map $\bm{R}$ is computed as follows:
\begin{align}
    \begin{split}
    \Bar{\bm{A}}_l&=\mathbb{E}_{h}[\nabla\bm{A}_l\odot\bm{A}_l], \;\nabla\bm{A}_l=\frac{\partial \bm{y}_t}{\partial \bm{A}_l}, \\
    \bm{R}&=\bm{R}+\Bar{\bm{A}}_l\cdot\bm{R},\quad\text{for }l\in\{1,2,\ldots,L\},
    \label{eq:relevancy_map_compute}
    \end{split}
\end{align}
where $\odot$ denotes the Hadamard product and $L$ is the total number of layers in the LVLM. In order to investigate the contribution of each layer to the final output, we visualize the image regions related to the output token through the visual relevancy map computed from each layer. Specifically, we consider the row of $\Bar{\bm{A}}_l\cdot\bm{R}$ corresponding to the output token. Then, we extract the visual token parts of the row to yield the visual relevancy map.

\begin{figure*}[t]
    \centering
    \begin{minipage}{\textwidth}
        \centering
        \includegraphics[width=0.85\linewidth]{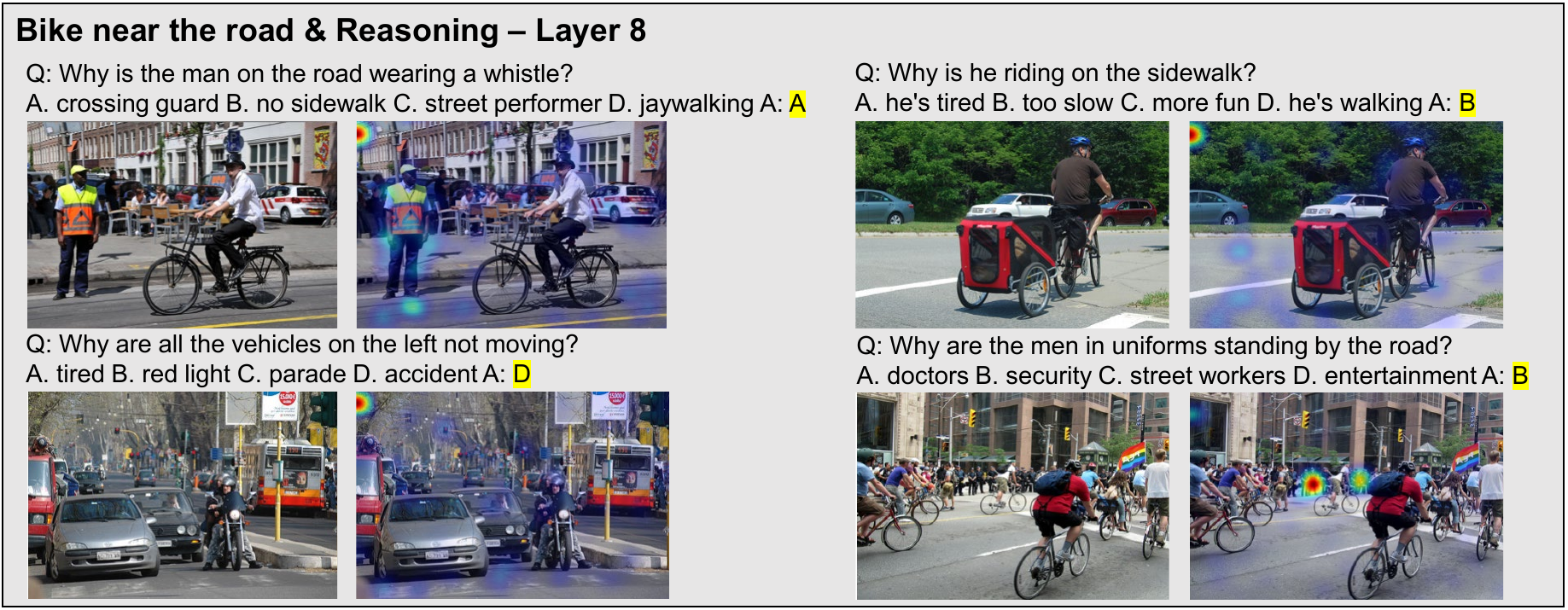}
    \end{minipage}
    \par
    \vspace{0.05in}
    \begin{minipage}{\textwidth}
        \centering
        \includegraphics[width=0.85\linewidth]{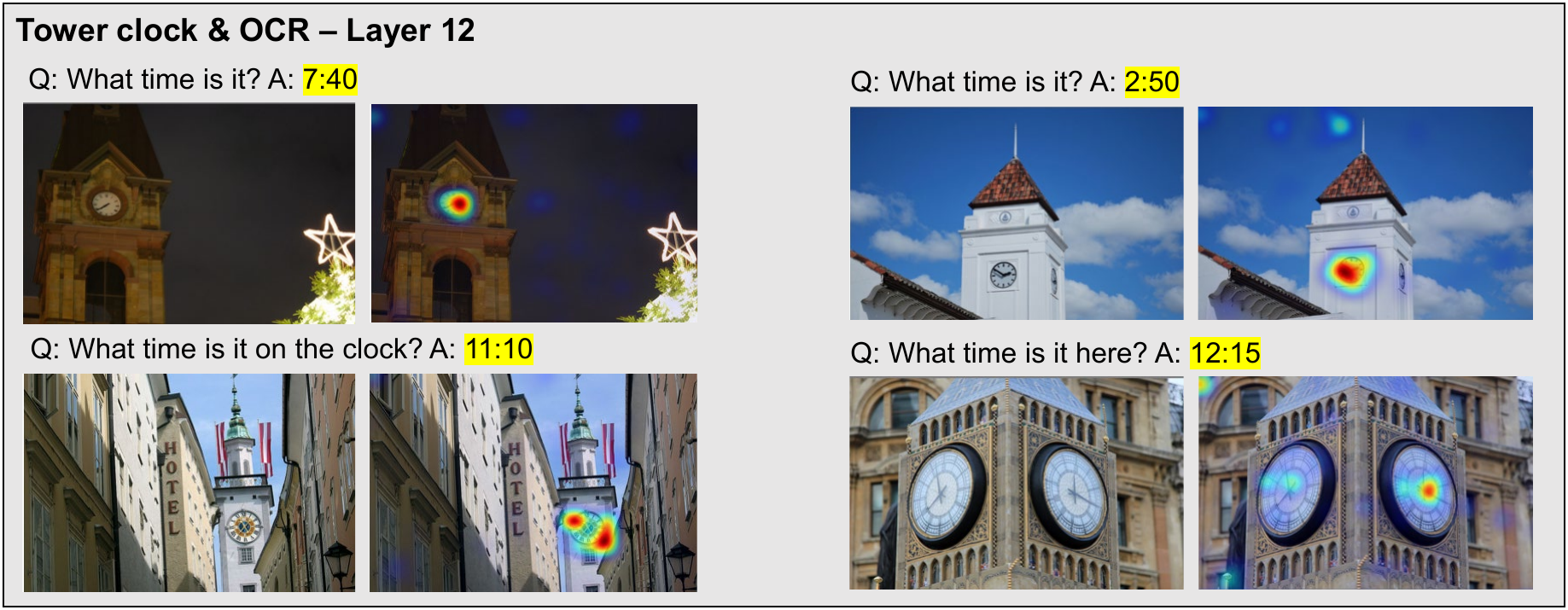}
    \end{minipage}
    \par
    \vspace{0.05in}
    \begin{minipage}{\textwidth}
        \centering
        \includegraphics[width=0.85\linewidth]{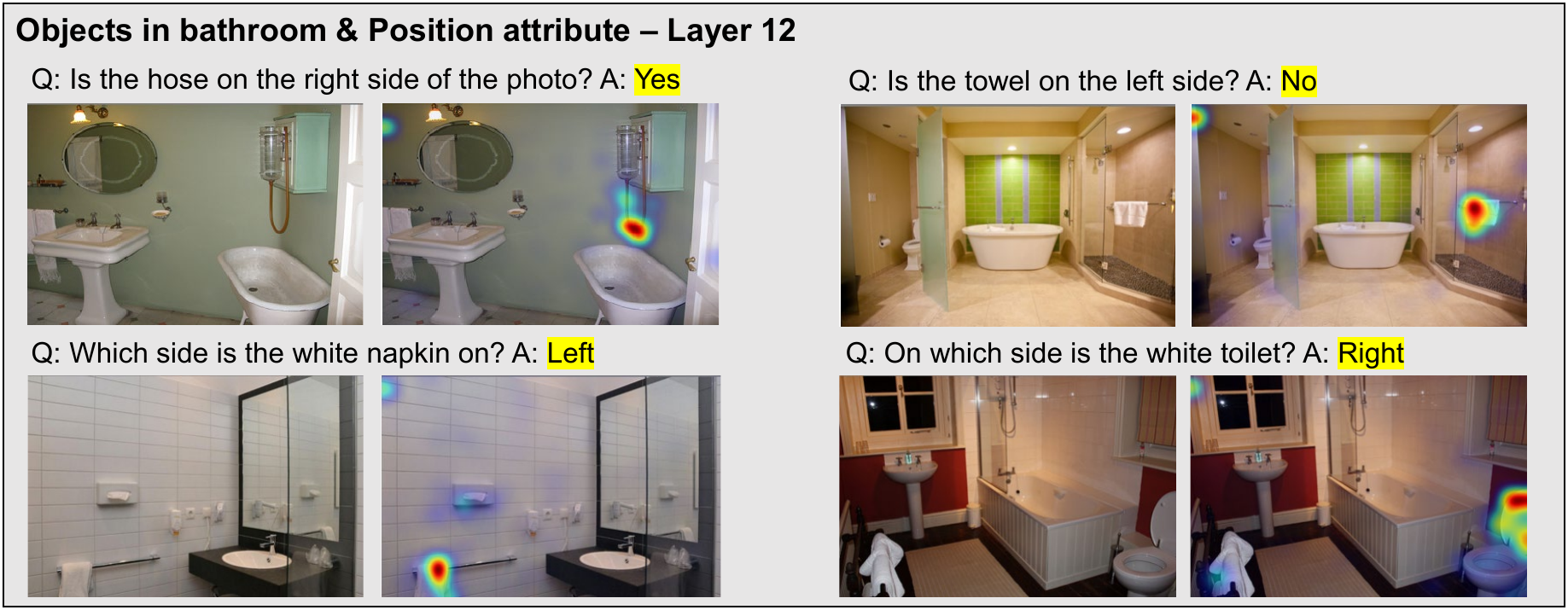}
    \end{minipage}
    \par
    \vspace{0.05in}
    \begin{minipage}{\textwidth}
        \centering
        \includegraphics[width=0.85\linewidth]{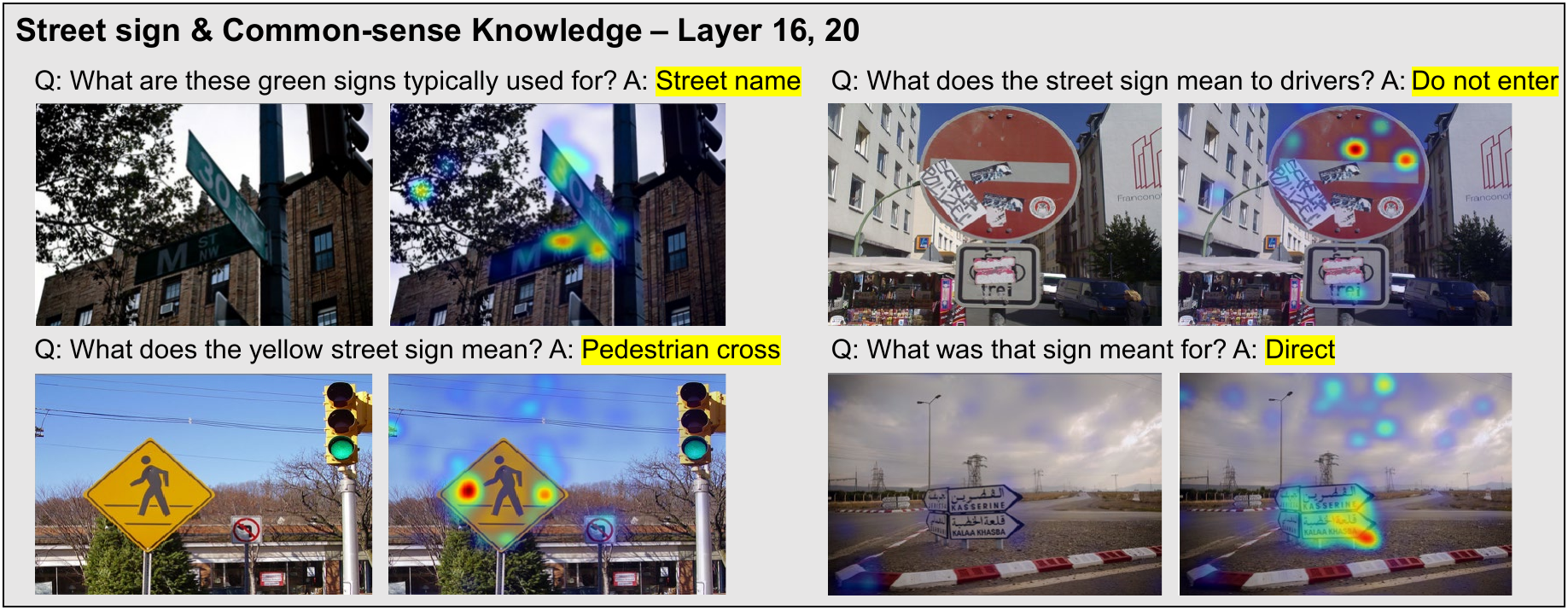}
    \end{minipage}
    \par
    \captionsetup{justification=justified}
    \caption{Relevancy maps visualization. We investigate which layer contributes most to the final output of the LVLM. This is done by visualizing relevancy maps of four samples from the same cluster. For each example, the left image is the original, while the right image shows the visualized relevancy map, highlighting regions most relevant to the LVLM output text colored in yellow. The top-left corner of each group explains the VL concept-skill composition and the layer number with the highest relevancy to the output.}
    \label{fig:supple_relevancy_visualization}
\end{figure*}

For the investigation, we inspect the 4th, 8th, 12th, 16th, and 20th layers of the TinyLLaVA-2B~\citep{Zhou2024tinyllava} model and identify the layer that activates the most relevant visual regions. The findings in~\Cref{fig:supple_relevancy_visualization} reveal that (1) the most relevant layer varies according to the concept-skill composition and (2) the most relevant layer is the same across diverse VIT data when the data shares a similar concept-skill composition. This supports our assumption that different layers contribute to distinct concepts and skills, allowing neuron activations from various layers to effectively group VIT data by their concept-skill composition. 

\section{Concept-Skill Clustering Visualization}
\label{appendix:cluster_visualize}
\begin{figure*}[t]
    \centering
    \begin{minipage}{\textwidth}
        \centering
        \includegraphics[width=0.85\linewidth]{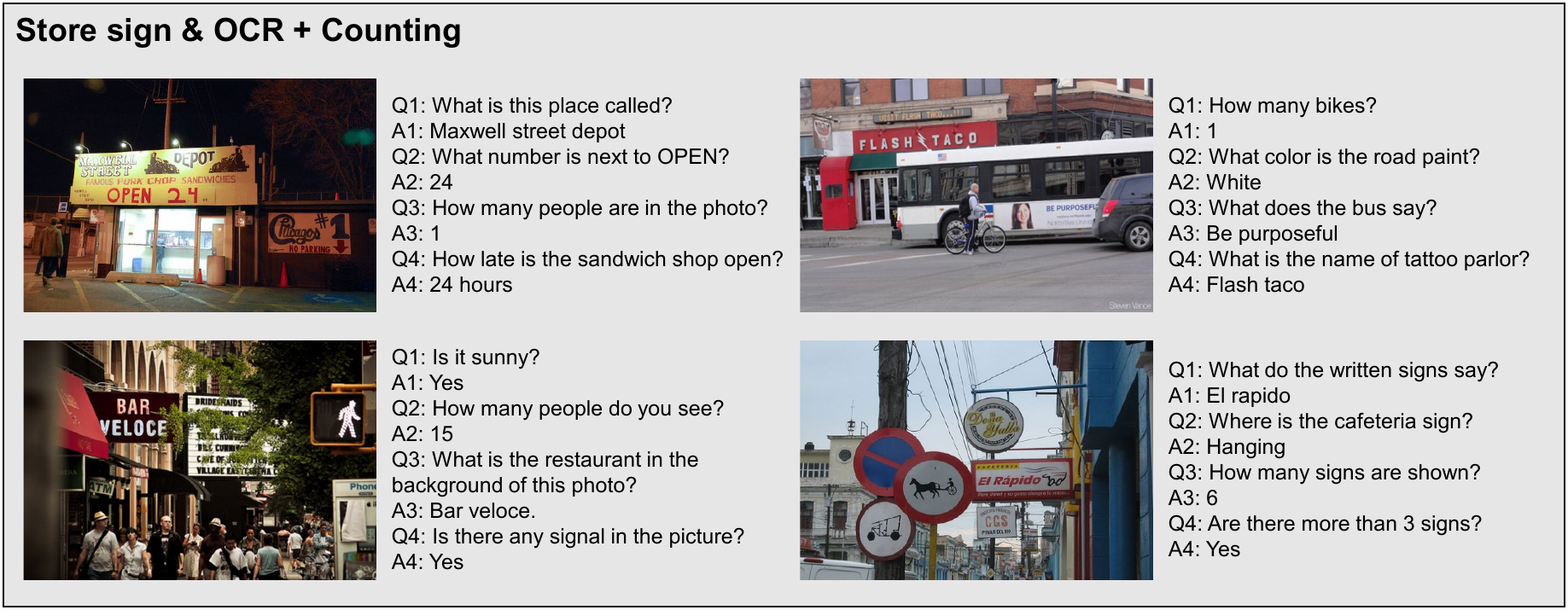}
    \end{minipage}
    \par
    \vspace{0.05in}
    \begin{minipage}{\textwidth}
        \centering
        \includegraphics[width=0.85\linewidth]{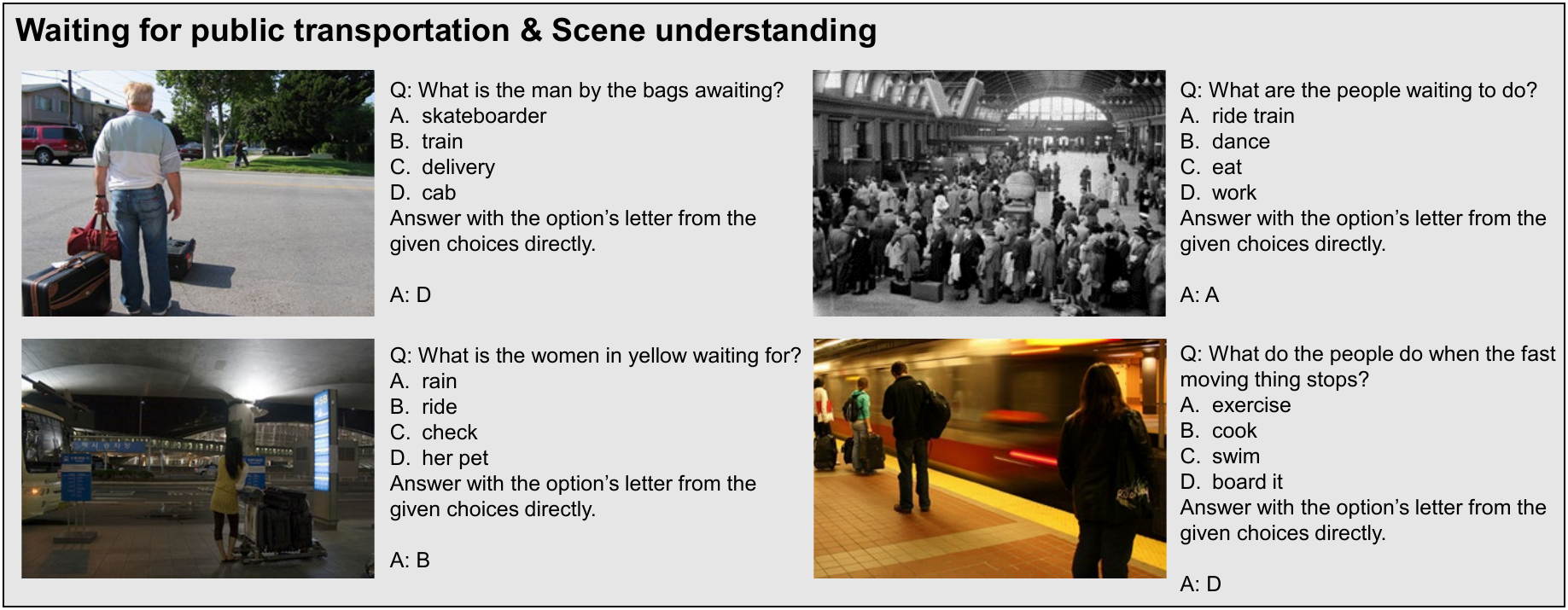}
    \end{minipage}
    \par
    \vspace{0.05in}
    \begin{minipage}{\textwidth}
        \centering
        \includegraphics[width=0.85\linewidth]{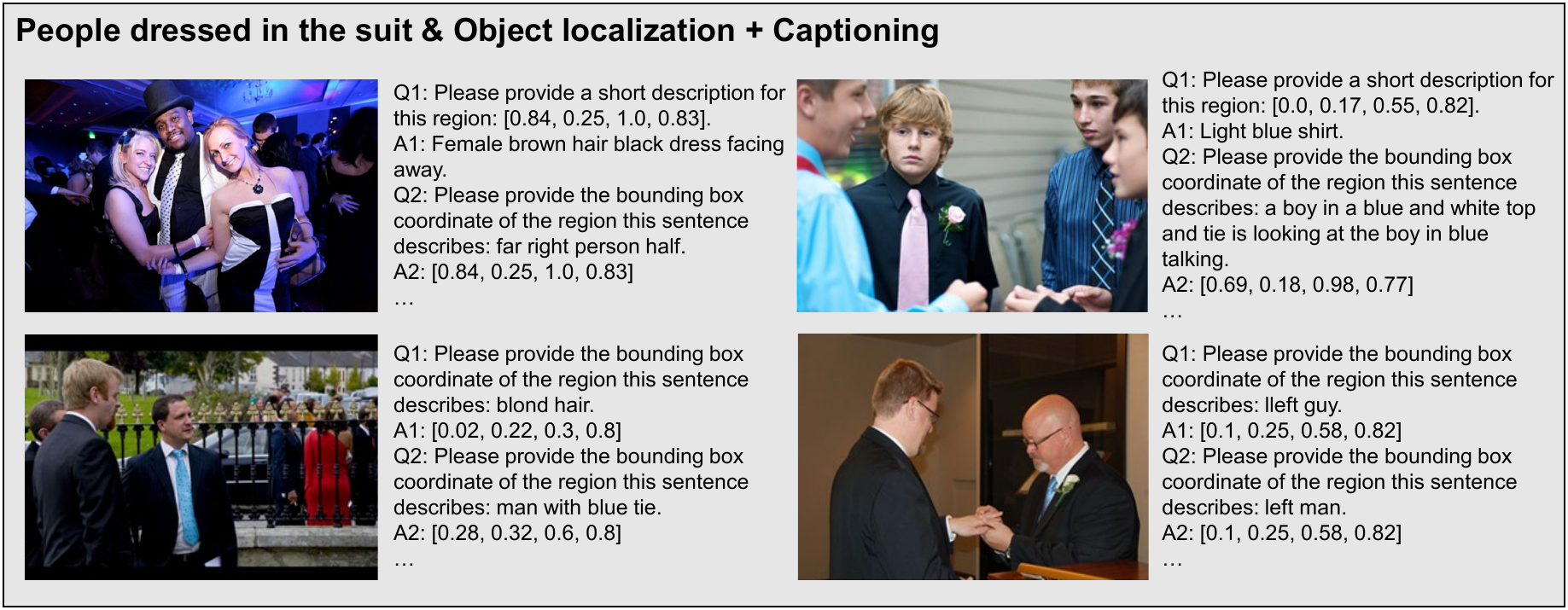}
    \end{minipage}
    \par
    \vspace{0.05in}
    \begin{minipage}{\textwidth}
        \centering
        \includegraphics[width=0.85\linewidth]{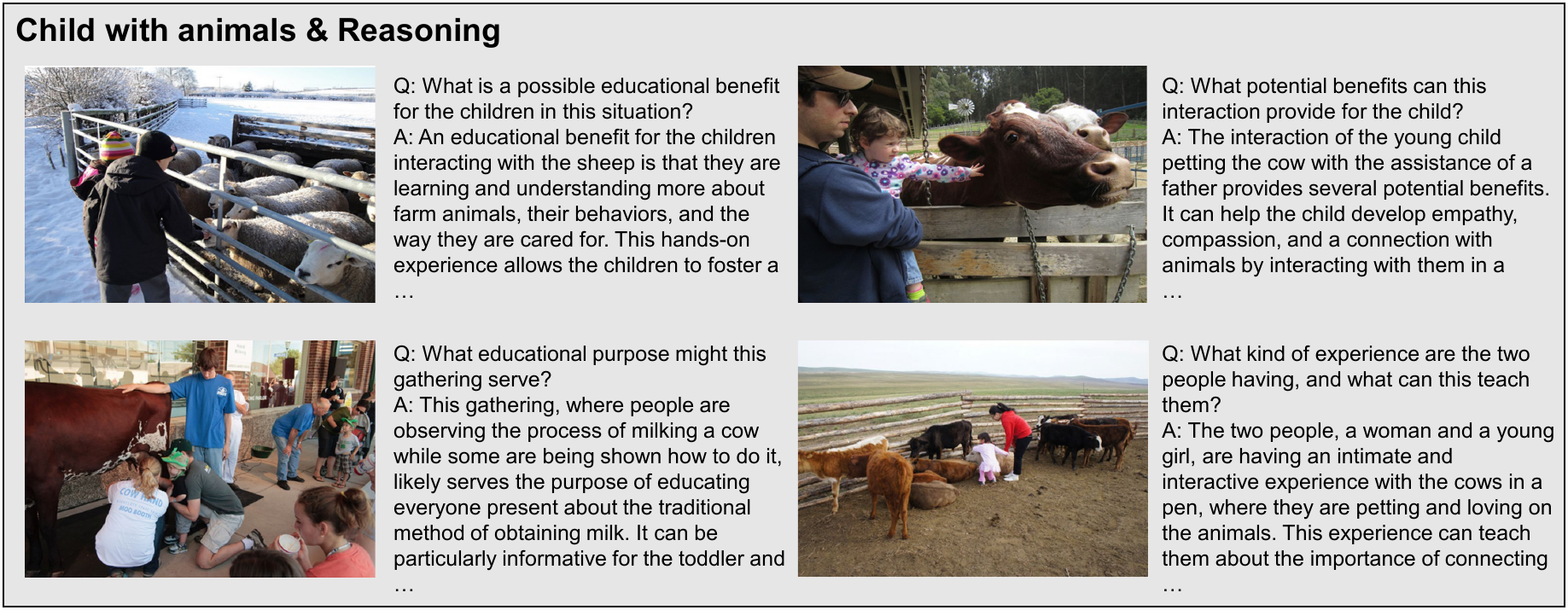}
    \end{minipage}
    \par
    \captionsetup{justification=justified}
    \caption{Examples of data clusters. We visualize four samples from the same cluster. The top-left corner of each group explains the VL concept-skill composition.}
    \label{fig:supple_cluster_visualization}
\end{figure*}

We visualize the clustering results of the gathered VIT data. The results are illustrated in~\Cref{fig:supple_cluster_visualization}. We observe that most clusters contain VIT data that encode similar concept-skill compositions. For instance, the first group in~\Cref{fig:supple_cluster_visualization} consists of samples requiring OCR and counting abilities to solve visual queries involving images with store signs. The second group features images of people waiting for public transportation and multiple-choice questions that require visual recognition and reasoning abilities. The third group shows a cluster of samples with images of people in suits and queries focusing on object localization and generating captions for given bounding boxes. Lastly, the bottom group includes images exhibiting children with animals and requiring the ability to reason about the educational benefits that the children might gain from interacting with the animals.

\section{In-Depth Analysis on Concept-Skill Composition Transferability \label{appendix:in_depth_concept_skill_analysis}}
\begin{table*}[t]
    \tiny
    \caption{Transferring to the larger target model. We validate if the coresets selected from TinyLLaVA-2B are transferable to LLaVA-1.5-13B finetining. We train the LLaVA-1.5-13B using coresets with 20\% sampling ratio and estimate performance on various multimodal benchmarks. The best and the second best results are highlighted in \textbf{bold} and \underline{underline}, respectively.}
    \centering
    \resizebox{\textwidth}{!}{
        \renewcommand{\arraystretch}{1.25}
        \renewcommand{\tabcolsep}{5.0pt}
        \begin{tabular}{l c c c c c c c cc c a}
             \toprule
             {\textbf{Method}} & {\textbf{VQAv2}} & {\textbf{GQA}} & {\textbf{VizWiz}} & {\textbf{SQA-I}} & {\textbf{TextVQA}} & {\textbf{POPE}} & {\textbf{MME}} & \multicolumn{2}{c}{\textbf{MMBench}} & {\textbf{LLaVA-}} & {\textbf{\relp(\%)}}\\
             & & & & & & & & {\textbf{en}} & {\textbf{cn}} & {\textbf{Wild}} & \\
             \midrule
             Full-Finetune &
             {\scriptsize 80.0} & {\scriptsize 63.3} & {\scriptsize 58.9} & {\scriptsize 71.2} & {\scriptsize 60.2} & {\scriptsize 86.7} & {\scriptsize 1541.7} & {\scriptsize 68.5} & {\scriptsize 61.5} & {\scriptsize 69.5} & {\scriptsize 100}\\
             \cmidrule{0-11}
              Random &
             {\scriptsize 76.7} & {\scriptsize \textbf{60.5}} & {\scriptsize 48.0} & {\scriptsize 68.8} & {\scriptsize \underline{57.7}} & {\scriptsize 84.8} & {\scriptsize 1484.9} & {\scriptsize 62.8} & {\scriptsize 55.2} & {\scriptsize 68.6} & {\scriptsize 94.0}\\
                CLIP-Score &
             {\scriptsize 75.3} & {\scriptsize 52.6} & {\scriptsize 42.2} & {\scriptsize 69.7} & {\scriptsize 57.3} & {\scriptsize \underline{85.4}} & {\scriptsize 1426.3} & {\scriptsize 60.4} & {\scriptsize 54.0} & {\scriptsize 68.1} & {\scriptsize 90.7}\\
              EL2N &
             {\scriptsize \underline{77.2}} & {\scriptsize 59.6} & {\scriptsize \textbf{54.8}} & {\scriptsize 69.9} & {\scriptsize 56.1} & {\scriptsize 84.1} & {\scriptsize \textbf{1531.0}} & {\scriptsize 59.3} & {\scriptsize 52.3} & {\scriptsize 65.8} & {\scriptsize 93.8}\\
              Perplexity &
             {\scriptsize 77.0} & {\scriptsize 58.5} & {\scriptsize 48.2} & {\scriptsize 68.7} & {\scriptsize 54.8} & {\scriptsize 83.1} & {\scriptsize 1508.8} & {\scriptsize 57.5} & {\scriptsize 50.3} & {\scriptsize \underline{68.7}} & {\scriptsize 91.6}\\
                SemDeDup &
             {\scriptsize 75.6} & {\scriptsize 57.5} & {\scriptsize 48.3} & {\scriptsize \textbf{70.5}} & {\scriptsize \underline{57.7}} & {\scriptsize 85.3} & {\scriptsize 1397.6} & {\scriptsize 59.0} & {\scriptsize 51.1} & {\scriptsize 68.7} & {\scriptsize 91.9}\\
                D2-Pruning &
             {\scriptsize 73.9} & {\scriptsize \textbf{60.5}} & {\scriptsize 49.8} & {\scriptsize \underline{70.4}} & {\scriptsize 55.2} & {\scriptsize 84.9} & {\scriptsize 1463.0} & {\scriptsize \textbf{67.3}} & {\scriptsize \textbf{59.9}} & {\scriptsize 66.5} & {\scriptsize \underline{94.7}}\\
                Self-Sup &
             {\scriptsize 76.3} & {\scriptsize \textbf{60.5}} & {\scriptsize 50.0} & {\scriptsize 70.2} & {\scriptsize 52.7} & {\scriptsize \underline{85.4}} & {\scriptsize 1463.8} & {\scriptsize 63.7} & {\scriptsize 57.6} & {\scriptsize 64.9} & {\scriptsize 93.6}\\
                Self-Filter &
             {\scriptsize 75.0} & {\scriptsize 59.8} & {\scriptsize 48.6} & {\scriptsize 69.5} & {\scriptsize 55.8} & {\scriptsize 84.5} & {\scriptsize 1446.9} & {\scriptsize 58.8} & {\scriptsize 51.8} & {\scriptsize \textbf{69.1}} & {\scriptsize 92.2}\\
                \cellcolor{gg}\ours (Ours) &
             \cellcolor{gg}{\scriptsize \textbf{77.8}} & \cellcolor{gg}{\scriptsize \underline{60.4}} & \cellcolor{gg}{\scriptsize \underline{51.6}} & \cellcolor{gg}{\scriptsize 70.0} & \cellcolor{gg}{\scriptsize \textbf{58.6}} & \cellcolor{gg}{\scriptsize \textbf{87.1}} & \cellcolor{gg}{\scriptsize \underline{1516.8}} & \cellcolor{gg}{\scriptsize \underline{64.0}} & \cellcolor{gg}{\scriptsize \underline{57.7}} & \cellcolor{gg}{\scriptsize 67.4} & \cellcolor{gg}{\scriptsize \textbf{95.9}}\\
             \bottomrule
        \end{tabular}
    }
    \label{tab:supple_llava_13b_eval}
\end{table*}

\subsection{Task-wise Transferability}
To further understand transferability, we calculate the transferability of LLaVA-1.5 tasks by averaging the cluster transferability of VIT data. We show the results in~\Cref{fig:avg_transfer}. We observe that VQA tasks, including VQAv2, GQA, OKVQA, and A-OKVQA, contain VIT data that transfers well to other data. In contrast, GPT-generated conversational tasks, including LLaVA-Conv, LLaVA-Detail, LLaVA-Rason, and ShareGPT, exhibit low transferability. This corresponds to the findings of \citet{Tiong2024lvlmfactoranal} that VQA tasks are effective for finetuning LVLMs. This alignment supports the efficacy of our approach in discovering the fine-grained concept-skill compositions and their transferability. We hypothesize that the high transferability of the VQA tasks is because these tasks mostly require abilities close to the fine-grained VL concepts and skills that can be shared with other tasks, as described in~\Cref{fig:shared_vl_concept_skill}, unlike more complex tasks.

\subsection{Concept-Skill with High Transferability \label{appendix:subsec:high_transfer_vis}}
In~\Cref{fig:supple_high_transfer_vis}, we visualize concept-skill compositions having the highest transferability for various VL task types. We define the VL task type of a cluster based on the task name associated with most of the cluster's data (e.g., VQAv2, GQA). Interestingly, GQA and LLaVA-Conv share a similar concept-skill composition as their most transferable concept-skill composition. This suggests that the transferability of VL concept-skill composition might be consistent across different VL tasks.

\subsection{Concept-Skill as Latent Factor of LVLM\label{appendix:subsec:shared_concept_skill}}
\begin{figure*}[t]
    \centering
    \begin{minipage}{\textwidth}
        \centering
        \includegraphics[width=0.85\linewidth]{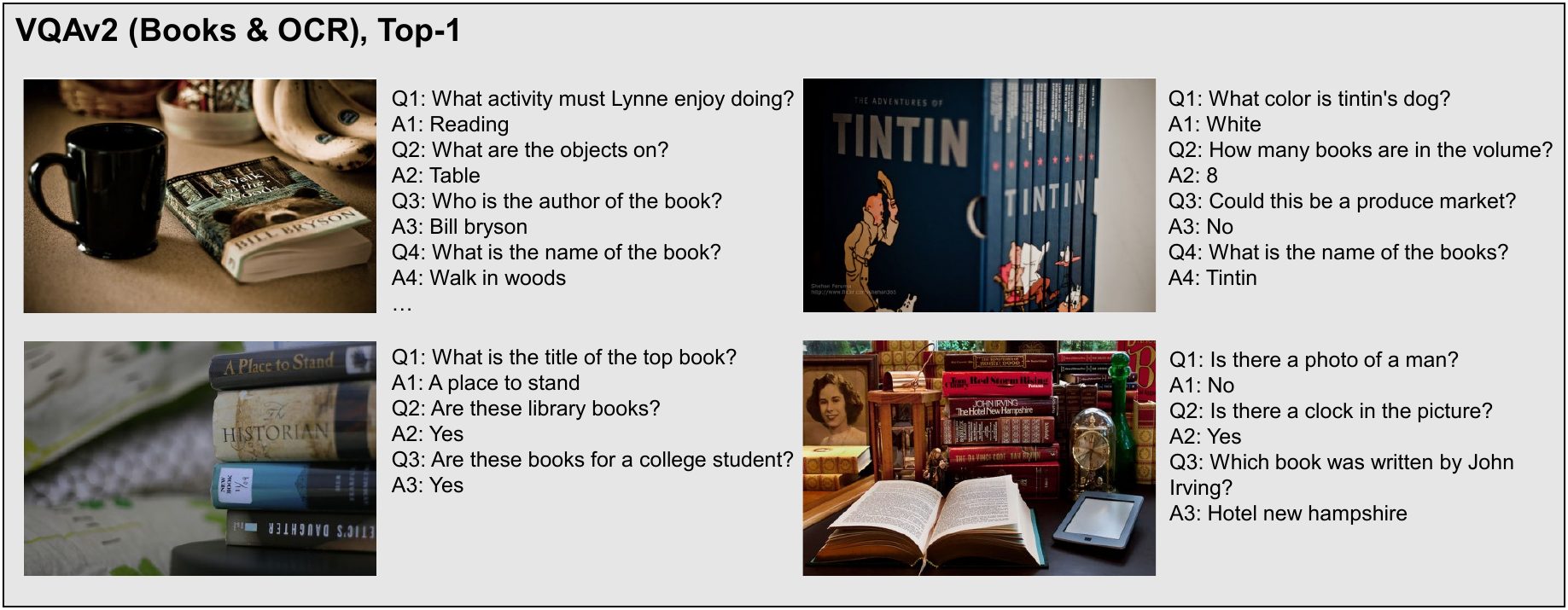}
    \end{minipage}
    \par
    \vspace{0.05in}
    \begin{minipage}{\textwidth}
        \centering
        \includegraphics[width=0.85\linewidth]{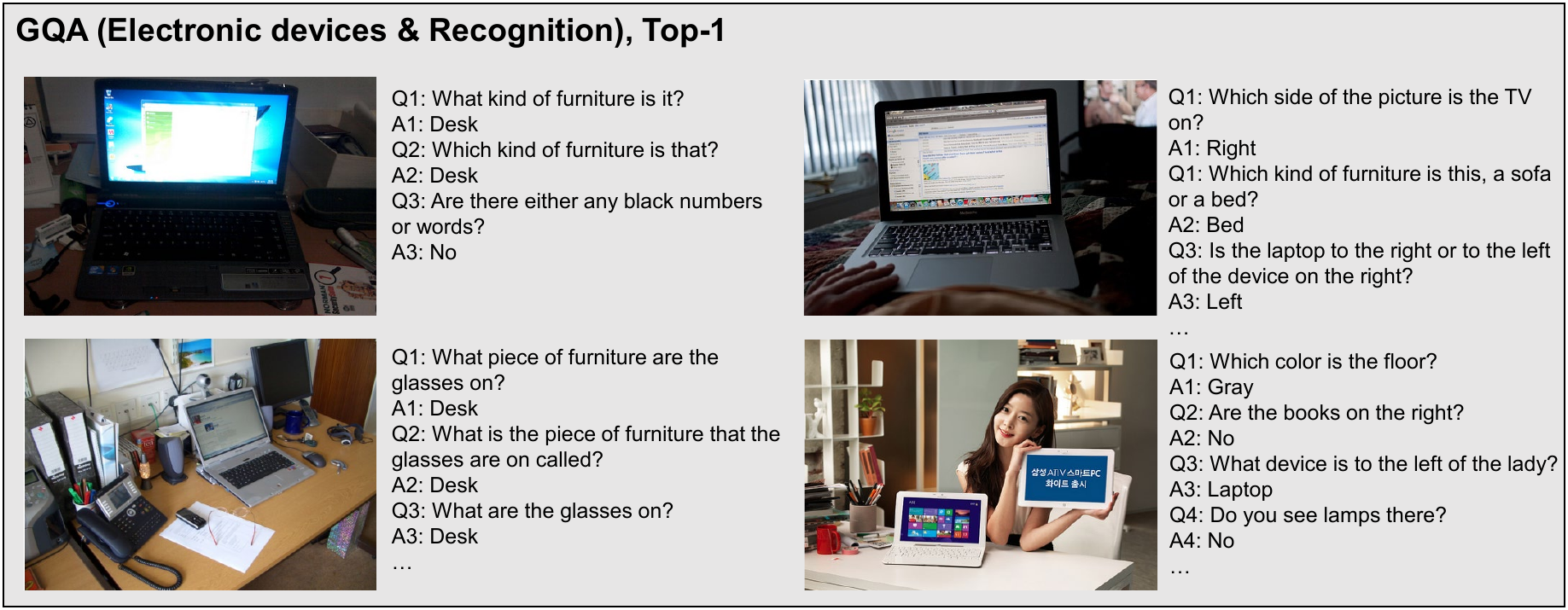}
    \end{minipage}
    \par
    \vspace{0.05in}
    \begin{minipage}{\textwidth}
        \centering
        \includegraphics[width=0.85\linewidth]{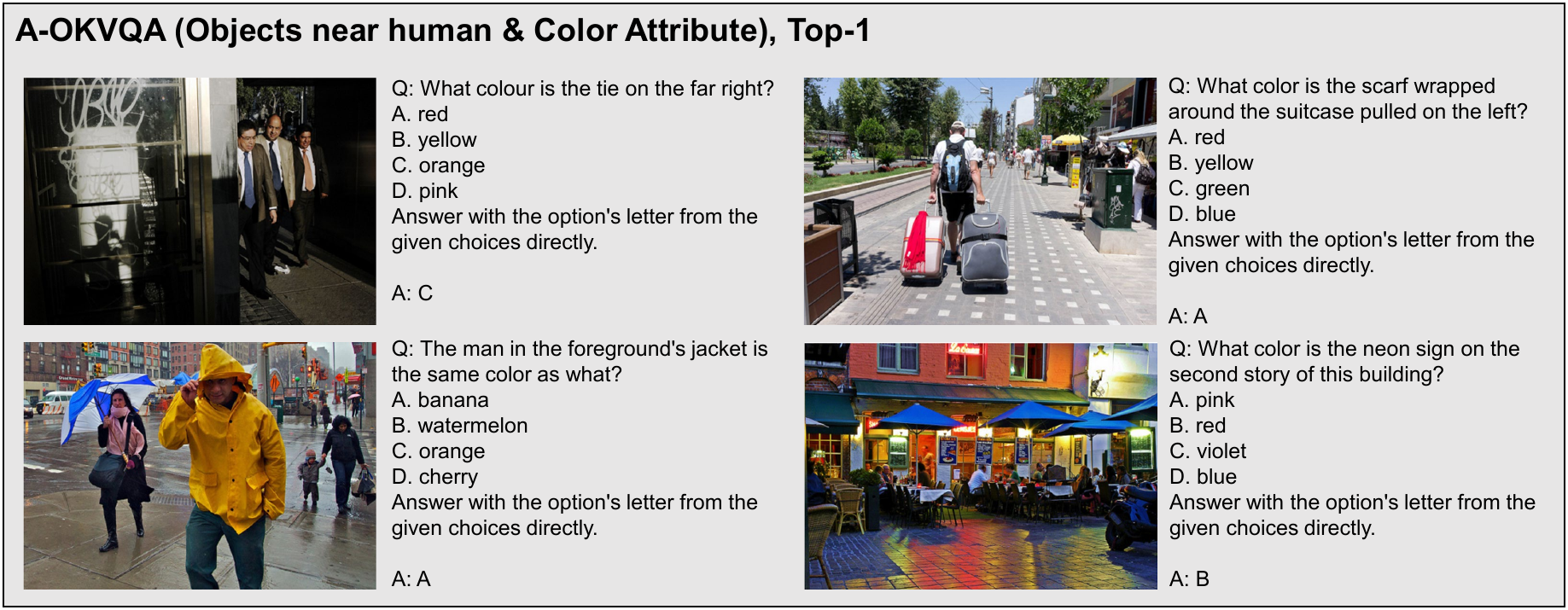}
    \end{minipage}
    \par
    \vspace{0.05in}
    \begin{minipage}{\textwidth}
        \centering
        \includegraphics[width=0.85\linewidth]{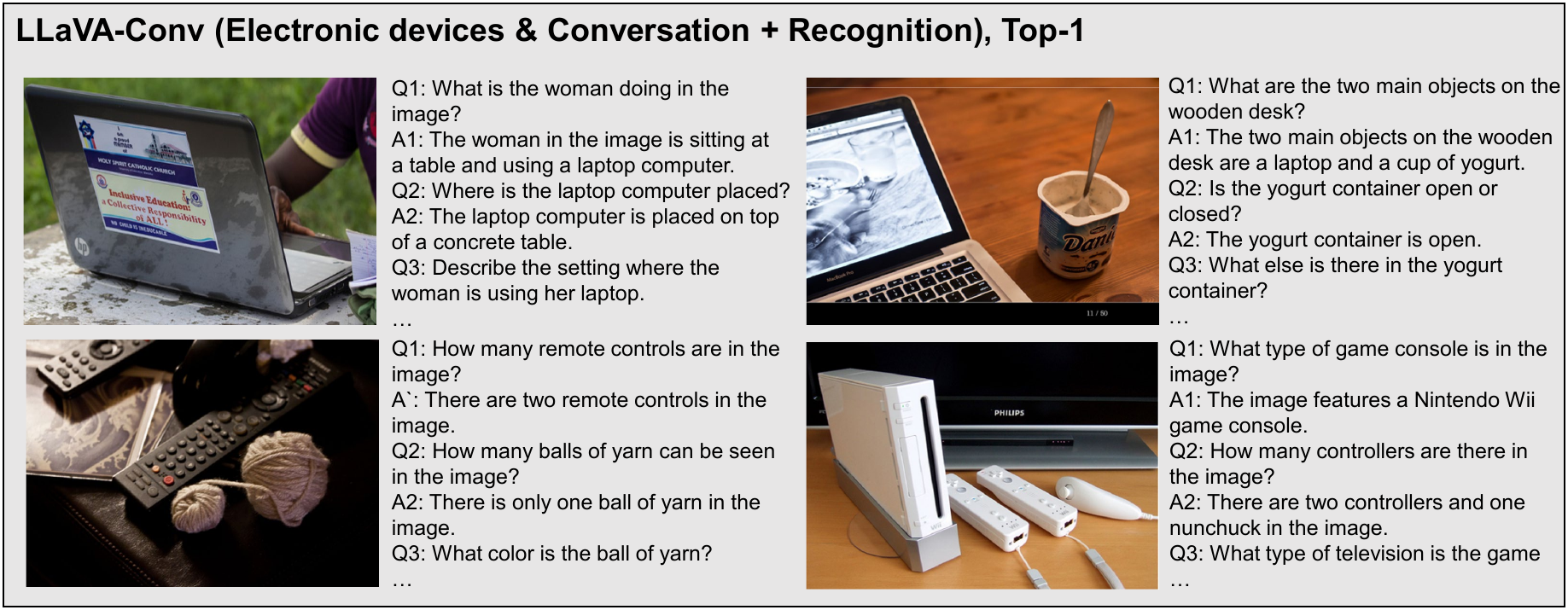}
    \end{minipage}
    \par
    \captionsetup{justification=justified}
    \caption{High transferability cluster sample visualization. We visualize the samples from the most transferable concept-skill composition for each VL task. The top-left corner of each group explains the VL task type and the VL concept-skill compositions. The VL task type for the group follows the task name where most of the data from the group are associated.}
    \label{fig:supple_high_transfer_vis}
\end{figure*}

We conduct an ablation study to verify if data clusters from different VL task types have high transferability with each other when they share a similar concept-skill composition. In this study, we select two clusters from different VL task types with a similar concept-skill composition (second and fourth groups in~\Cref{fig:supple_high_transfer_vis}), using the first cluster as the source and the second cluster as the target. Additionally, we employ 49 randomly selected source clusters and measure transferability from the source clusters to the target cluster (Eq.~\ref{eq:eval_gain}). The source cluster, sharing a similar concept-skill composition with the target, ranks in the top 5 of the 50 source clusters in terms of test loss gain, exhibiting high transferability to the target cluster. This suggests that concept-skill compositions resemble fine-grained latent factors that constitute LVLM abilities. Thus, these fine-grained VL concepts and skills must be considered to effectively reduce data redundancy and build a well-generalized LVLM.

\section{Concept-Skill Diversity within Coresets\label{appendix:diversity_analysis}}
\begin{figure*}[t]
    \centering
    \begin{minipage}{\textwidth}
        \centering
        \includegraphics[width=0.85\linewidth]{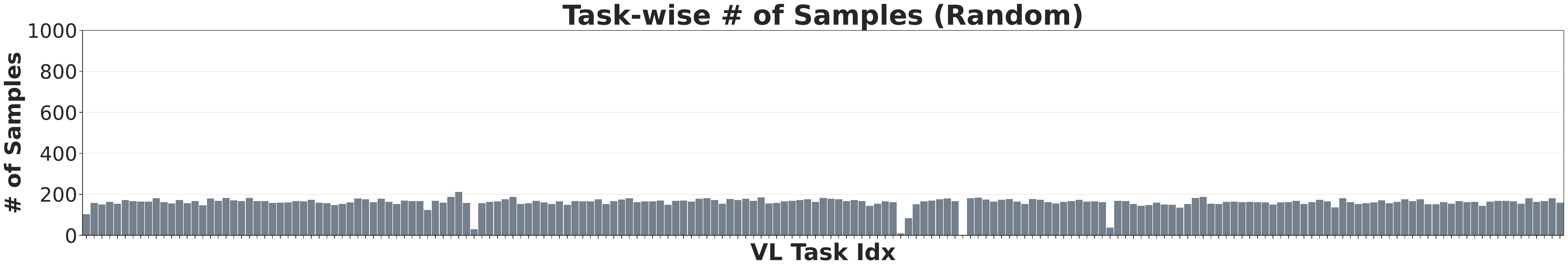}
    \end{minipage}
    \par
    \vspace{0.05in}
    \begin{minipage}{\textwidth}
        \centering
        \includegraphics[width=0.85\linewidth]{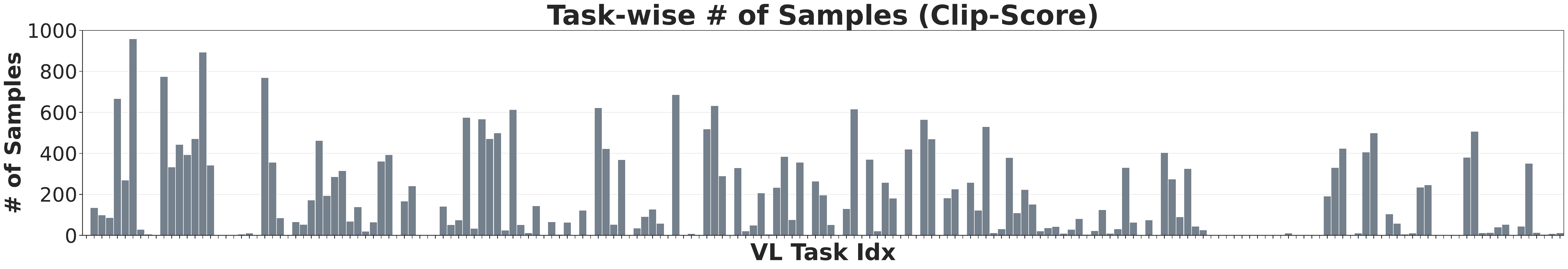}
    \end{minipage}
    \par
    \vspace{0.05in}
    \begin{minipage}{\textwidth}
        \centering
        \includegraphics[width=0.85\linewidth]{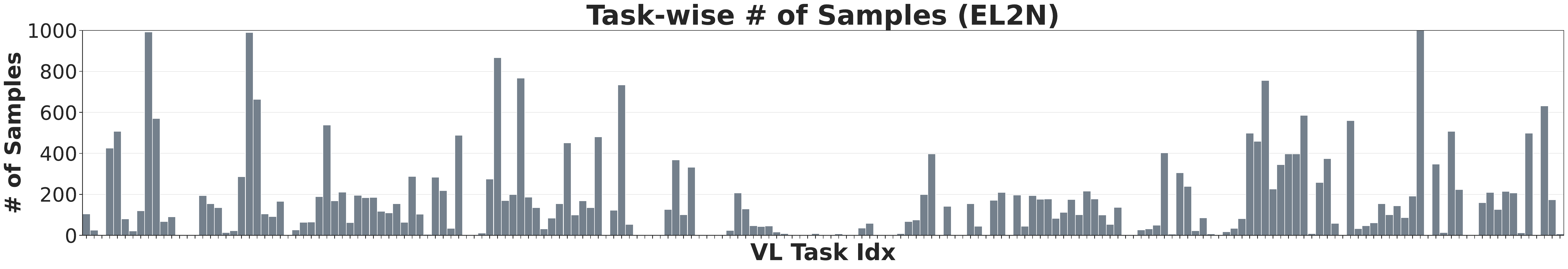}
    \end{minipage}
    \par
    \vspace{0.05in}
    \begin{minipage}{\textwidth}
        \centering
        \includegraphics[width=0.85\linewidth]{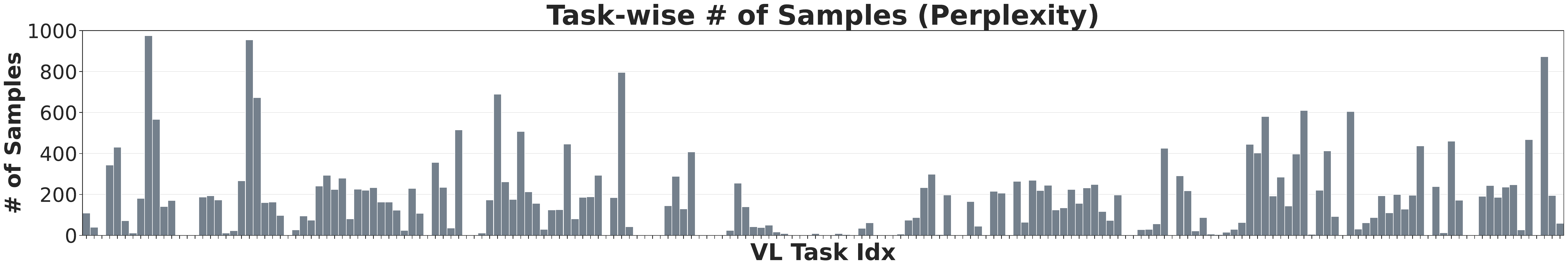}
    \end{minipage}
    \par
    \vspace{0.05in}
    \begin{minipage}{\textwidth}
        \centering
        \includegraphics[width=0.85\linewidth]{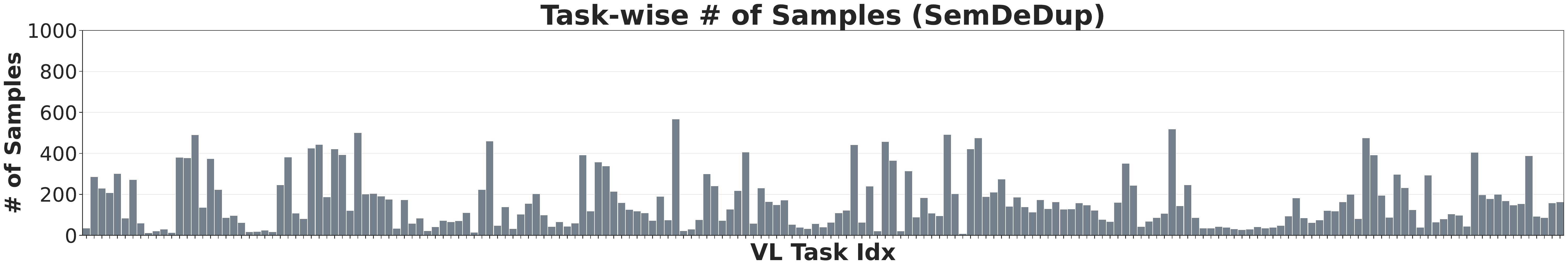}
    \end{minipage}
    \par
    \vspace{0.05in}
    \begin{minipage}{\textwidth}
        \centering
        \includegraphics[width=0.85\linewidth]{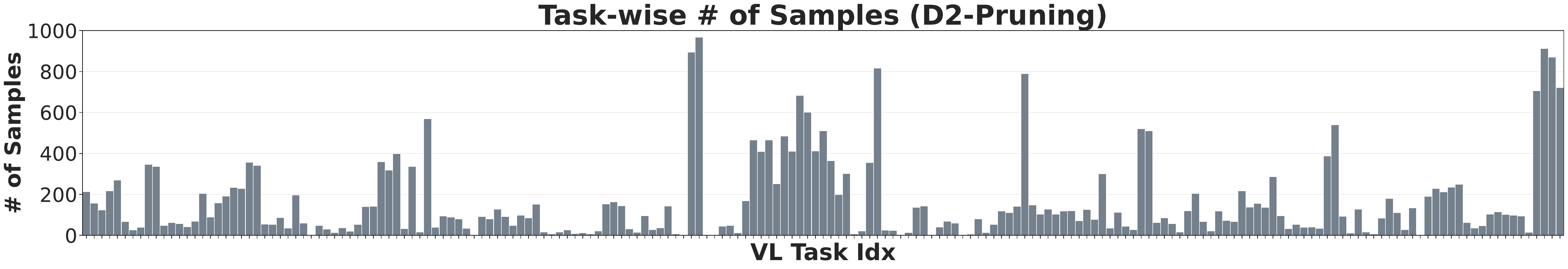}
    \end{minipage}
    \par
    \vspace{0.05in}
    \begin{minipage}{\textwidth}
        \centering
        \includegraphics[width=0.85\linewidth]{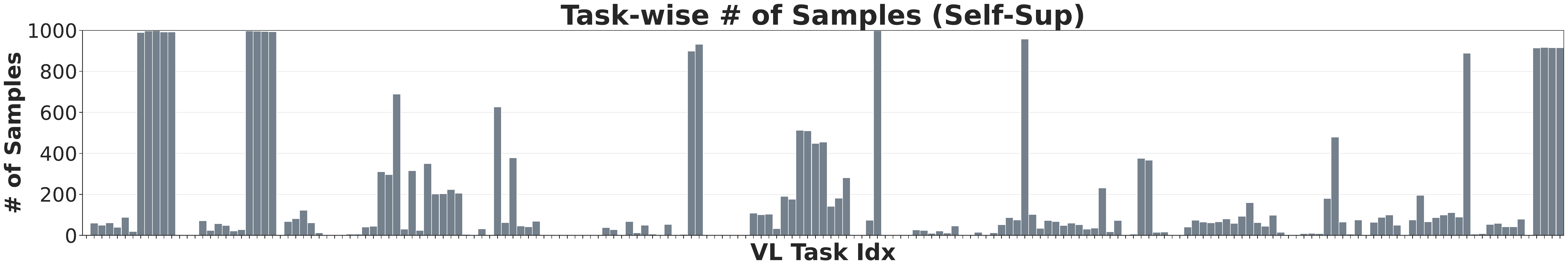}
    \end{minipage}
    \par
    \vspace{0.05in}
    \begin{minipage}{\textwidth}
        \centering
        \includegraphics[width=0.85\linewidth]{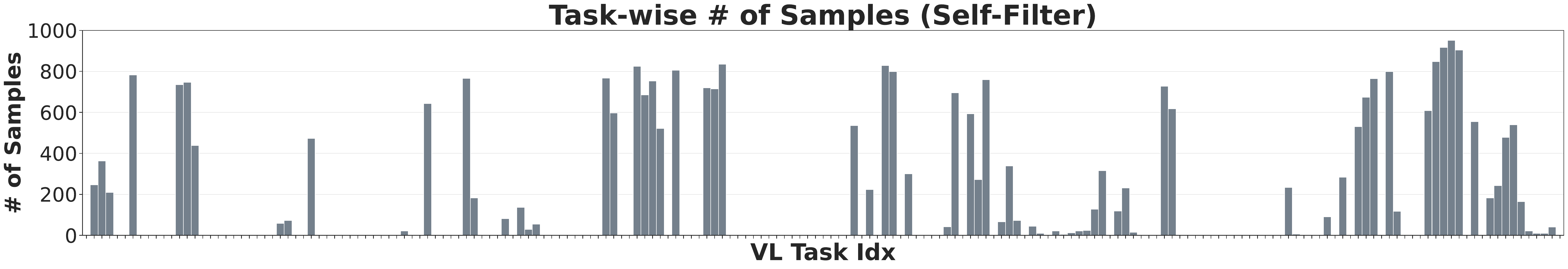}
    \end{minipage}
    \par
    \vspace{0.05in}
    \begin{minipage}{\textwidth}
        \centering
        \includegraphics[width=0.85\linewidth]{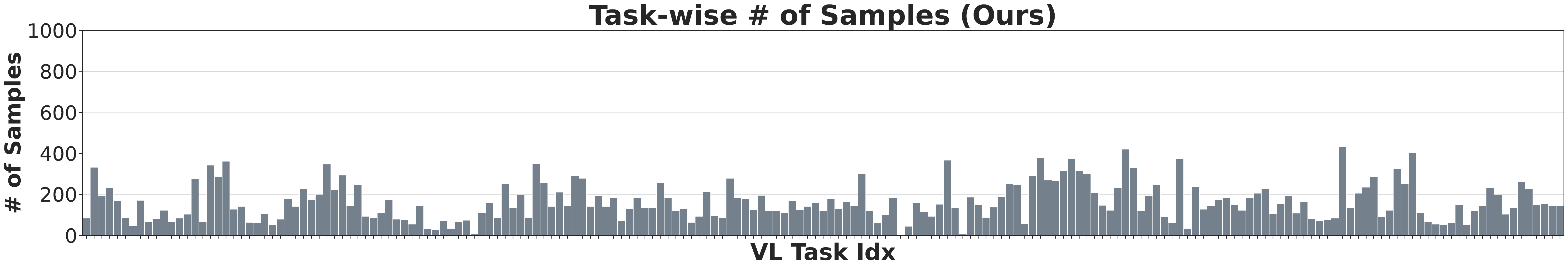}
    \end{minipage}
    \par
    \vspace{0.05in}
    \captionsetup{justification=justified}
    \caption{The number of selected samples per VL task in the Vision-Flan VIT dataset. The horizontal axis denotes the VL task index in the dataset, and the vertical axis denotes the number of samples. Baseline methods result in biased coresets. In contrast, our method achieves a more balanced sample selection across diverse tasks, leading to better LVLM generalization.}
    \label{fig:supple_vision_flan_diversity}
\end{figure*}

Our method selects data from various clusters to ensure a high diversity of VL concept-skill compositions within the coreset. To demonstrate the efficacy of our method, we compare the diversity within the coreset by our method with those by the baseline methods. Specifically, we use the 191 tasks from the Vision-Flan dataset as proxies for different concept-skill compositions, as there are no ground-truth compositions. We then count the number of selected samples for each task. The results, summarized in~\Cref{fig:supple_vision_flan_diversity}, indicate that baseline methods select most data from only a few tasks, leading to biased selection and undermining LVLM generalization. This bias explains why most baselines perform worse than random sampling in our experiments. In contrast, our method achieves a more balanced selection across the various tasks.
\begin{figure*}[t]
    \centering
    \begin{minipage}{0.68\linewidth}
        \captionsetup{type=figure}
            \centering
            \captionof{figure}{Hyperparameter search. We examine the effect of the temperature ($\tau$) and the number of clusters ($K$).}
            \vspace{-0.05in}
            \includegraphics[width=0.49\linewidth]{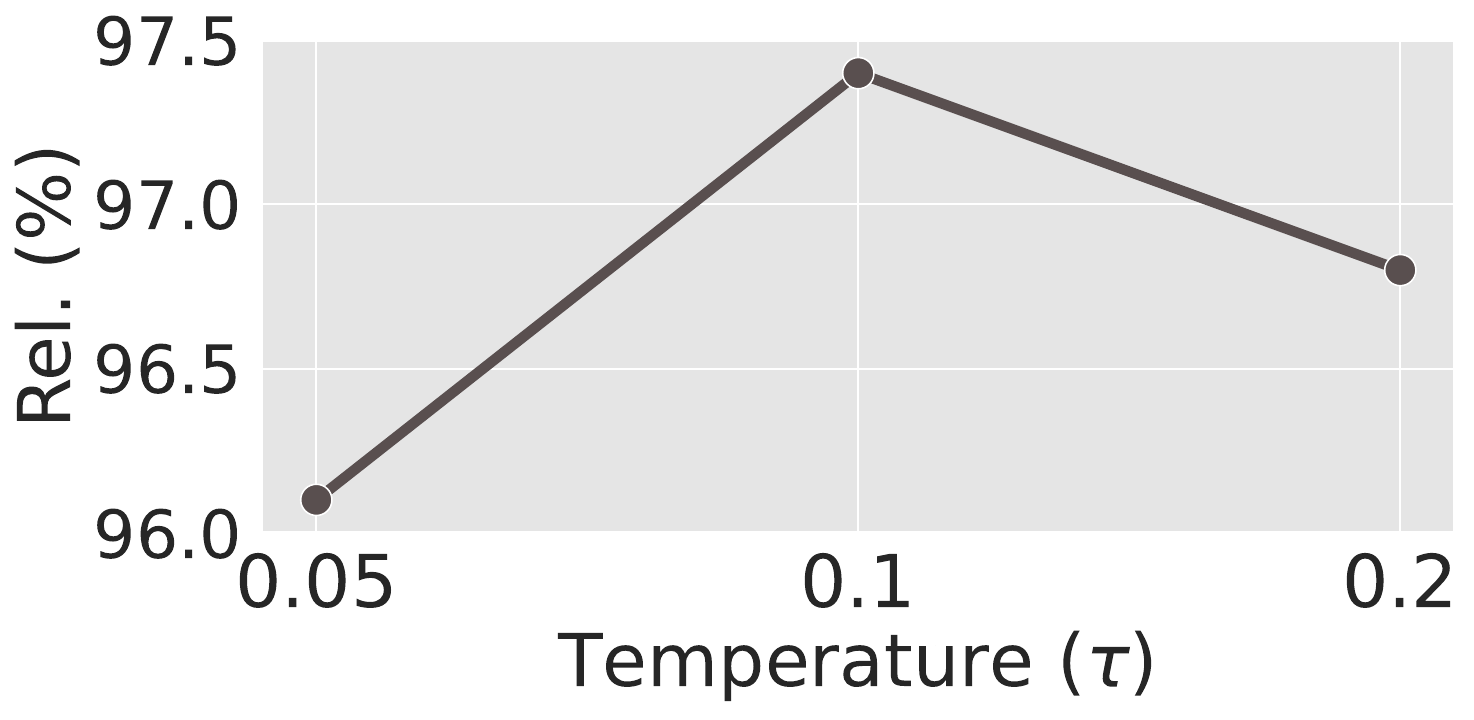}
            \includegraphics[width=0.49\linewidth]{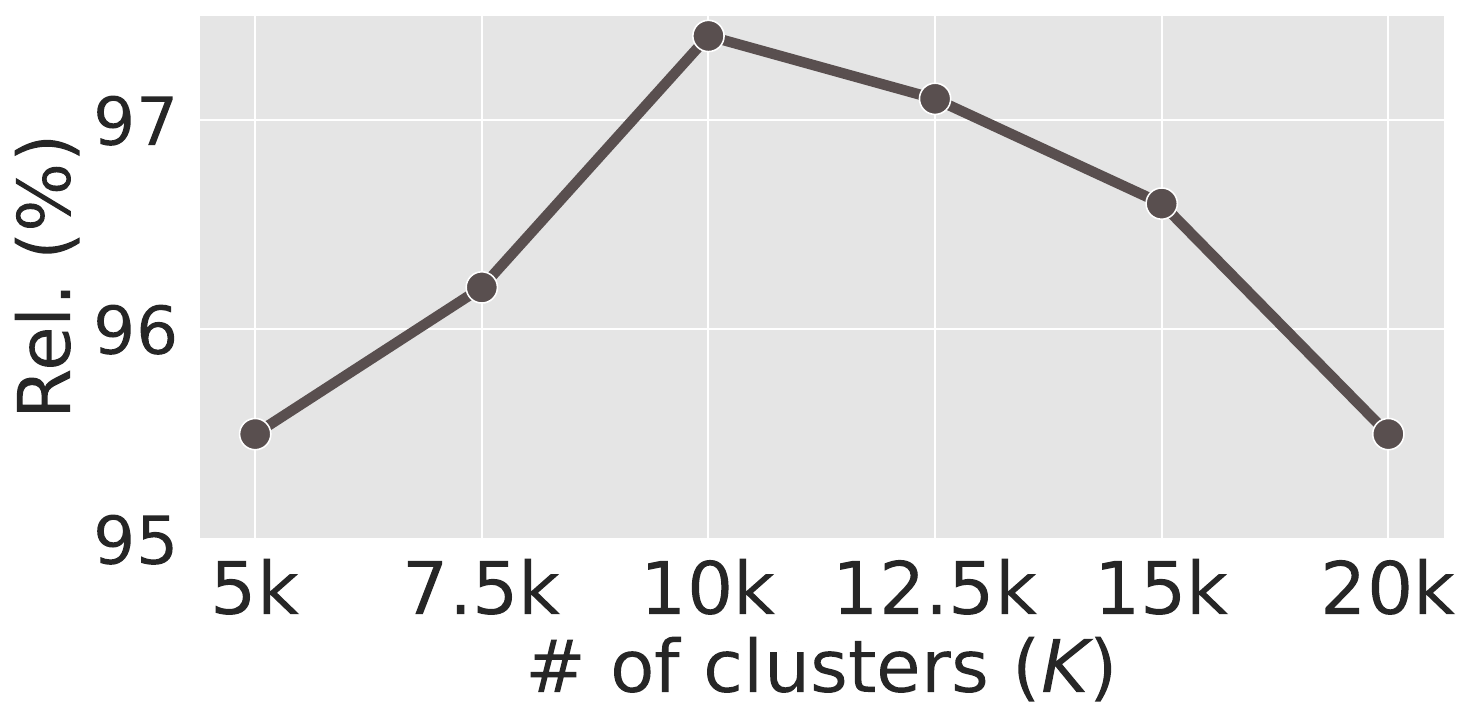}
            \label{fig:effect_hyperparameter}
    \end{minipage}
    \hspace{0.05in}
    \begin{minipage}{0.28\linewidth}
        \captionsetup{type=table}
            \centering
            \captionof{table}{We investigate the impact of various representations of multimodal neuron activation.}
            \resizebox{\linewidth}{!}{
            \renewcommand{\arraystretch}{1.}
            \renewcommand{\tabcolsep}{5pt}
            \begin{tabular}{l c}
                 \toprule
                 {\textbf{Neuron Activation}} & \relp (\%)\\
                 \midrule
                 Boolean & 95.7\\
                 Last layer & 96.5\\
                 MSA layers & \textbf{97.4}\\
                 FFN layers & 96.0\\
                 \bottomrule
            \end{tabular}
            }
            \label{tab:neuron_activation}
    \end{minipage}
\end{figure*}

\begin{table}[t]
    \tiny
    \caption{Impact of a reference model training dataset. We use TinyLLaVA-2B finetuned on the LLaVA-1.5 dataset as a reference model to collect coresets from the Vision-Flan dataset with 16.7\% sampling ratio. The best and the second best results are highlighted in \textbf{bold} and \underline{underline}, respectively.}
    \centering
    \resizebox{\linewidth}{!}{
        \renewcommand{\arraystretch}{1.3}
        \renewcommand{\tabcolsep}{1.0pt}
        \begin{tabular}{l c c c c c a}
             \toprule
             {\textbf{Method}} & {\textbf{MMBench-en}} & {\textbf{MME}} & {\textbf{MM-Vet}} & {\textbf{POPE}} & {\textbf{SQA-I}} & {\textbf{\relp (\%)}}\\
             \midrule
             Full-Finetune &
             {\scriptsize 53.4} & {\scriptsize 1287.5} & {\scriptsize 25.6} & {\scriptsize 84.2} & {\scriptsize 61.3} & {\scriptsize 100}\\
             \cmidrule{0-6}
              EL2N &
             {\scriptsize 41.8} & {\scriptsize 1082.0} & {\scriptsize 23.9} & {\scriptsize 82.6} & {\scriptsize 61.7} & {\scriptsize 90.9}\\
              Perplexity &
             {\scriptsize 45.7} & {\scriptsize 1001.7} & {\scriptsize 26.1} & {\scriptsize \underline{81.9}} & {\scriptsize \textbf{64.8}} & {\scriptsize 93.7}\\
                SemDeDup &
             {\scriptsize 46.8} & {\scriptsize 1129.7} & {\scriptsize \textbf{27.2}} & {\scriptsize 82.5} & {\scriptsize 64.3} & {\scriptsize 96.9}\\
                D2-Pruning &
             {\scriptsize \underline{48.1}} & {\scriptsize \textbf{1143.0}} & {\scriptsize \underline{27.0}} & {\scriptsize \underline{83.4}} & {\scriptsize 63.1} & {\scriptsize \underline{97.3}}\\
                Self-Sup &
             {\scriptsize 47.1} & {\scriptsize 1084.6} & {\scriptsize 23.5} & {\scriptsize 81.7} & {\scriptsize 63.5} & {\scriptsize 93}\\
                \cellcolor{gg} \ours (Ours) &
             \cellcolor{gg}{\scriptsize \textbf{51.7}} & \cellcolor{gg}{\scriptsize \underline{1139.0}} & \cellcolor{gg}{\scriptsize 26.9} & \cellcolor{gg}{\scriptsize \textbf{84.0}} & \cellcolor{gg}{\scriptsize \underline{64.5}} & \cellcolor{gg}{\scriptsize \textbf{99.1}}\\
             \bottomrule
        \end{tabular}
    }
    \label{tab:vision_flan_cross_valid}
\end{table}

\section{Additional Experimental Results}

\subsection{Transfering to Larger Target Model \label{appendix:subsec:transfer_larger}}
We evaluate the performance of the larger target model (LLaVA-1.5-13B) finetuned on coresets gathered by the small LVLM (TinyLLaVA-2B). \Cref{tab:supple_llava_13b_eval} summarizes the performances across various benchmarks. The results demonstrate the effectiveness of our method in selecting a coreset that can be successfully transferred to the larger target model.

\begin{figure*}[ht!]
\begin{minipage}{\textwidth}
    \centering
    \begin{algorithm}[H]
        \caption{\ours Data Selection Algorithm}
        \label{alg:selection_algo}
        \begin{algorithmic}[1]
            \REQUIRE $K$: the number of clusters, $N_{\text{core}}$: target coreset size
            \STATE Extract multimodal neuron activations $\bm{u}^{m}$ from the full dataset. \COMMENTARR{Eq.~\ref{eq:multi_neuron_act}}
            \STATE Cluster $\bm{u}^{m}$ into $K$ clusters to form a set of clusters $\mathcal{C}\!=\!\{\mathcal{C}_1,\mathcal{C}_2,\ldots,\mathcal{C}_{K}\}$.
            \STATE Compute cluster transferability $S_{i}\!=\!\mathbb{E}_{j}\left(\texttt{cos}(\bm{e}_i,\bm{e}_j)\right)$,\;\;$i\!\in\!\{1,2,\ldots,K\}$ \COMMENTARR{Eq.~\ref{eq:cosine_sim}}
            \STATE Compute cluster density $D_{i}\!=\!\mathbb{E}_{p,q\sim C_i}\left(d(p,q)\right)$,\;\;$i\!\in\!\{1,2,\ldots,K\}$ \COMMENTARR{Eq.~\ref{eq:density_cal}}
            \STATE Calculate cluster categorical distribution $P_i \propto \exp(S_i / ( \tau D_i))$.
            \FOR{$i = 1,2,\ldots,K$}
                \STATE $i$-th cluster empty coreset $\mathcal{C}_{i}^{\prime}$.
                \STATE $i$-th cluster target sample size $N_{\text{core},i}\!=\! N_{\text{core}}P_{i}$.
                \WHILE{$|\mathcal{C}_{i}^{\prime}| < N_{\text{core},i}$}
                    \STATE $k\!=\!\underset{j\!\in\!\mathcal{C}_{i}\setminus\mathcal{C}_{i}^{\prime}}{\text{argmin}}\;\text{MMD}^{2}\left(\mathcal{C}_{i},\mathcal{C}_{i}^{\prime}\cup \{j\}\right)$ \COMMENTARR{Eq.~\ref{eq:mmd_cal}}
                    \STATE $\mathcal{C}_{i}^{\prime}\leftarrow\mathcal{C}_{i}^{\prime}\cup\{k\}$
                \ENDWHILE
            \ENDFOR
            \RETURN $\mathcal{C}_{1}^{\prime}\cup\mathcal{C}_{2}^{\prime}\cup\ldots\cup\mathcal{C}_{K}^{\prime}$
        \end{algorithmic}
    \end{algorithm}
\end{minipage}
\vspace{-0.1in}
\end{figure*}
\subsection{Robustness of Reference Model \label{appendix:subsec:robust_reference}}
We investigate the robustness of our method when the reference model is finetuned on a VIT dataset different from a target VIT dataset. To this end, we use the TinyLLaVA-2B finetuned on the LLaVA-1.5 VIT dataset, to perform coreset selection from the Vision-Flan dataset. The results are summarized in~\Cref{tab:vision_flan_cross_valid}. \ours continues to show performance comparable to full-finetuning while outperforming other baseline methods.

\subsection{Hyperparameters}
We conduct ablation studies on hyperparameters of our method, which include the number of clusters ($\bm{K}$) and the temperature ($\bm{\tau}$). The results, summarized in~\Cref{fig:effect_hyperparameter}, reveal that a sufficiently large number of clusters is essential to ensure cluster purity and diversity of VL concept-skill compositions, ensuring effective representation of the compositions and enhancing the generalization ability of LVLM. Furthermore, we find that setting the temperature too low leads to a biased coreset selection, as most samples are then selected from a few clusters. This undermines the diversity within the coreset, leading to a decline in overall performance.

\subsection{Multimodal Neuron Activation} 
We further analyze the impact of different multimodal neuron activations on the performance of our method. \ours selects neuron activations from the MSA blocks across the 4th, 8th, 12th, 16th, and 20th layers of the reference model. We experiment with different neuron activations and present the results in~\Cref{tab:neuron_activation}. Transforming the neuron activations from the MSA blocks into boolean vectors by mapping negative values to -1 and positive values to 1 causes a significant performance drop, likely due to substantial information loss, yielding inaccurate clustering and transferability calculation. Extracting neuron activations only from the last layer of the reference model causes a slight performance decrease. As discussed in~\Cref{sec:subsec:clustering}, LVLM abilities stem from various layers. Hence, relying on the last layer captures only a small portion of these capabilities, leading to the performance decline. Finally, utilizing the neuron activations from the MSA blocks gives superior performance compared to using activations from the FFN blocks. We believe this is because MSA layers use self-attention to share multimodal information, providing richer multimodal understanding.

\section{The \ours Algorithm \label{appendix:algorithm}}
In~\Cref{alg:selection_algo}, we outline our VIT data selection procedure, which involves several key stages: clustering the data (lines 1-2), calculating the cluster categorical distribution (lines 3-5), and selecting samples from each cluster (lines 6-15).

\clearpage

\end{document}